\documentclass[runningheads]{llncs}

\usepackage[mobile]{eccv}

\usepackage{eccvabbrv}
\usepackage{graphicx}
\usepackage{booktabs}
\usepackage{pifont}
\usepackage{bbding}
\usepackage{rotating}
\usepackage[accsupp]{axessibility}  
\usepackage{multirow}
\usepackage[table,xcdraw]{xcolor}

\usepackage[pagebackref,breaklinks,colorlinks]{hyperref}

\usepackage{orcidlink}

\usepackage[table]{xcolor}
\definecolor{customyellow}{HTML}{FFFFC7}
\newcommand{\highlight}[1]{\colorbox{customyellow}{#1}}

\begin{document}

\title{GGRt: Towards Pose-free Generalizable 3D Gaussian Splatting in Real-time} 

\titlerunning{Abbreviated paper title}

\author{Hao Li\inst{1,\thanks{Equal Contribution},\thanks{Work done interning at Baidu VIS}} \and
Yuanyuan Gao\inst{1, \footnotemark[1]} \and
Chenming Wu\inst{2, \footnotemark[1]} \and
Dingwen Zhang\inst{1,\thanks{Corresponding Author}} \and \\
Yalun Dai\inst{3} \and 
Chen Zhao\inst{2} \and
Haocheng Feng\inst{2} \and
Errui Ding\inst{2} \and \\
Jingdong Wang\inst{2} \and
Junwei Han\inst{1}
}

\authorrunning{H. Li et al.}

\institute{Brain and Artificial Intelligence Lab, Northwestern Polytechnical University \email{lifugan\_10027@outlook.com,zhangdingwen2006yyy@gmail.com \\ yyg7645@gmail.com, junweihan2010@gmail.com} \and
Department of Computer Vision, Baidu Inc.
\email{\{wuchenming, zhaochen03, fenghaocheng, dingerrui, wangjingdong\}@baidu.com}\and
Nanyang Technological University \email{daiy0018@e.ntu.edu.sg}}

\maketitle

\begin{abstract}
This paper presents GGRt, a novel approach to generalizable novel view synthesis that alleviates the need for real camera poses, complexity in processing high-resolution images, and lengthy optimization processes, thus facilitating stronger applicability of 3D Gaussian Splatting (3D-GS) in real-world scenarios. Specifically, we design a novel joint learning framework that consists of an Iterative Pose Optimization Network (IPO-Net) and a Generalizable 3D-Gaussians (G-3DG) model. With the joint learning mechanism, the proposed framework can inherently estimate robust relative pose information from the image observations and thus primarily alleviate the requirement of real camera poses. Moreover, we implement a deferred back-propagation mechanism that enables high-resolution training and inference, overcoming the resolution constraints of previous methods. To enhance the speed and efficiency, we further introduce a progressive Gaussian cache module that dynamically adjusts during training and inference. As the first pose-free generalizable 3D-GS framework, GGRt achieves inference at $\ge$ 5 FPS and real-time rendering at $\ge$ 100 FPS. Through extensive experimentation, we demonstrate that our method outperforms existing NeRF-based pose-free techniques in terms of inference speed and effectiveness. It can also approach the real pose-based 3D-GS methods. Our contributions provide a significant leap forward for the integration of computer vision and computer graphics into practical applications, offering state-of-the-art results on LLFF, KITTI, and Waymo Open datasets and enabling real-time rendering for immersive experiences. Project page: \href{https://3d-aigc.github.io/GGRt}{https://3d-aigc.github.io/GGRt}.


\end{abstract}

\section{Introduction}
\label{sec:intro}

Recently invented Neural Radiance Fields (NeRF)~\cite{mildenhall2021nerf} and 3D Gaussian Splatting (3D-GS)~\cite{kerbl20233d} bridge the gap between computer vision and computer graphics in the tasks of image-based novel view synthesis and 3D reconstruction. With a variety of follow-up variants, they are rapidly pushing the boundary towards revolutionizing many areas, such as virtual reality, film production, immersive entertainment, etc. 
To enhance generalization capabilities across previously unseen scenes, recent developments have introduced innovative approaches such as the generalizable NeRF~\cite{wang2022attention} and 3D-GS~\cite{charatan2023pixelsplat}.

\begin{figure}
    \centering
    \includegraphics[width=0.93\linewidth]{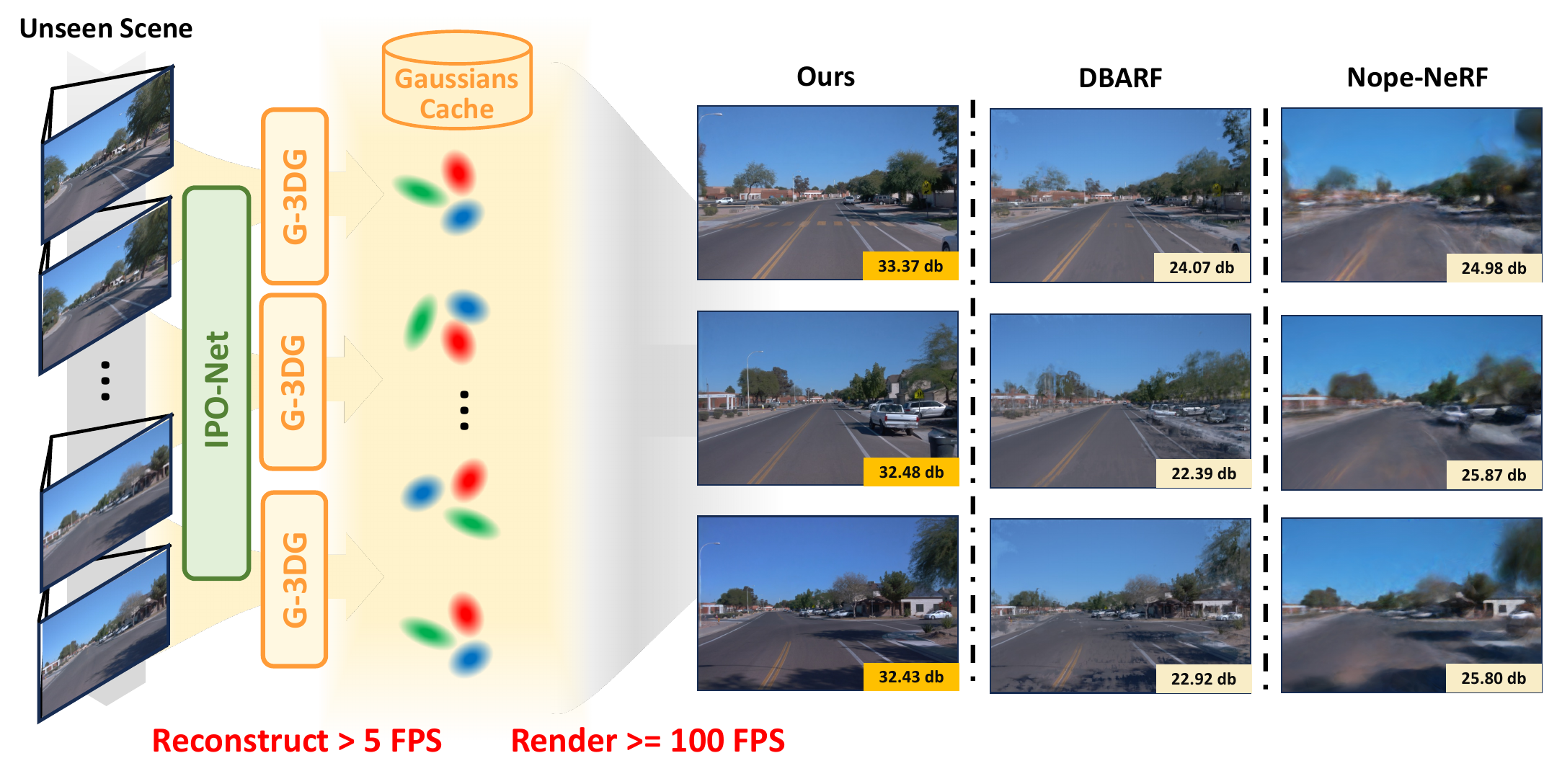}
    \caption{Our proposed GGRt stands for the first pose-free generalizable 3D Gaussian splatting approach, capable of inference at over 5 FPS, and delivering real-time rendering performance.}
    \label{fig:teaser}
\vspace{-0.5cm}
\end{figure}
Despite their ability to reconstruct new scenes without optimization, the previous works usually rely on the actual camera pose for each image observation, which actually cannot always be captured accurately in real-world scenarios. Besides, these methods show unsatisfactory view synthesis performance and struggle to reconstruct at higher resolutions due to the large number of parameters used. Last but not least, for such methods, each time when synthesizing a novel view demands a complete forward pass of the whole network, making real-time rendering intractable. 


To tackle these challenges, this paper proposes GGRt, which brings the benefits of a primitive-based 3D representation—fast and memory-efficient rendering—to the generalizable novel view synthesis under the pose-free condition. Specifically, we introduce a novel pipeline that jointly learns the IPO-Net and the G-3DG model. Such a pipeline can estimate relative camera pose information robustly and thus effectively alleviate the requirement for real camera poses.
Subsequently, we develop a deferred back-propagation (DBP) mechanism, allowing our method to efficiently perform high-resolution training and inference, a capability that surpasses the low-resolution limitations of existing methods \cite{lai2021video,sajjadi2023rust,coponerf,tian2023mononerf}.
Furthermore, we also design an innovative Gaussians cache module with the idea of reusing the relative pose information and image features of the reference views in two continuous training and inferencing iterations. Thus, the Gaussians cache can progressively grow and diminish throughout the training and inferencing processes, further accelerating the speed of both. 

To the best of our knowledge, our work stands for the first pose-free generalizable 3D Gaussian splatting, inference at $\ge$ 5 FPS, and rendering in real-time at $\ge$ 100 FPS.
Extensive experiments demonstrate that our method surpasses existing NeRF-based pose-free approaches in inference speed and effectiveness. Compared to pose-based 3D-GS methods, our approach provides faster inference and competitive performance, even without the camera pose prior. 

\section{Related Work}
\label{sec:related_work}

\subsection{Generalizable Novel View Synthesis}
Pioneering approaches involving novel view synthesis leverage image-based rendering techniques, such as light field rendering~\cite{suhail2022light,sitzmann2021light} and view interpolation~\cite{wang2022attention,yao2018mvsnet}. The introduction of NeRF~\cite{mildenhall2021nerf} marks a significant milestone that uses neural networks to model the volume scene function and demonstrates impressive results in this task but requires per-scene optimization and accurate camera poses. To address the problem of generalization, researchers have explored several directions. For instance, PixelNeRF~\cite{yu2021pixelnerf} presents a NeRF architecture that is conditioned on image inputs in a fully convolutional fashion. 
NeuRay~\cite{liu2022neural} enhances the NeRF framework by predicting the visibility of 3D points relative to input views, allowing the radiance field construction to concentrate on visible image features. Furthermore, GNT~\cite{wang2022attention} integrates multi-view geometry into an attention-based representation, which is then decoded through an attention mechanism in the view transformer for rendering novel views.

A recent work LRM~\cite{hong2023lrm} and its multi-view version~\cite{li2023instant3d}, also adopt a transformer for generalizable scene reconstruction using either a single image or posed four images. However, those works only demonstrate the capability in object-centric scenes, while our work targets a more ambitious goal of being generalizable in both indoor and outdoor scenes. Fu et al. \cite{fu20233d} propose to use a generalizable neural field from posed RGB images and depth maps, eschewing a fusion module. Our work, in contrast, requires only camera input without pose information. 

The aforementioned works use implicit representation inherited from NeRF and its variants, showing slow training and inferencing speed. 
Differently, pixleSplat~\cite{charatan2023pixelsplat} is the first generalizable 3D-GS work that tackles the problem of synthesizing novel views between a pair of images. However, it still requires accurate poses and only supports a pair of images as inputs. Instead, our work dismisses the demand for image poses and supports large-scale scene inference with unlimited images as reference views.

\subsection{Pose-free Modeling for Novel View Synthesis}
The first attempt towards pose-free novel view synthesis is iNeRF~\cite{yen2021inerf}, which uses key-point matching to predict camera poses. NeRF--~\cite{wang2021nerf} proposes to optimize camera pose embeddings and NeRF jointly. \cite{lin2021barf} proposes to learn neural 3D representations and register camera frames using coarse-to-fine positional encodings. \cite{bian2023nope} integrates scale and shift-corrected monocular depth priors to train their model, enabling the joint acquisition of relative poses between successive frames and novel view synthesis of the scenes. \cite{meuleman2023progressively} employs a strategy that synergizes pre-trained depth and optical-flow priors. This approach is used to progressively refine blockwise NeRFs, facilitating the frame-by-frame recovery of camera poses. 

The implicit modeling inherent to NeRF complicates the simultaneous optimization of scene and camera poses. However, the recent innovation of 3D-GS provides an explicit point-based scene representation, enabling real-time rendering and highly efficient optimization. A recent work~\cite{fu2023colmap} pushes the boundary of simultaneous scene and pose optimization. However, those approaches need tremendous efforts in training and optimization per scene.

In generalizable settings, SRT~\cite{srt22}, VideoAE~\cite{lai2021video}, RUST~\cite{sajjadi2023rust}, MonoNeRF~\cite{tian2023mononerf}, DBARF~\cite{chen2023dbarf} and FlowCam~\cite{smith2023flowcam} learn a generalizable scene representation from unposed videos using NeRF's implicit representation. Those works show unsatisfactory view synthesis performance without per-scene optimization and inherent all the drawbacks NeRF originally had, such as real-time rendering of explicit primitives. PF-LRM~\cite{wang2023pf} extends LRM to be applicable in pose-free scenes by using a differentiable PnP solver, but it shows the same limitations of LRM~\cite{hong2023lrm} mentioned above. To the best of our knowledge, our work stands for the first pose-free generalizable 3D-GS that enables efficient inferencing and real-time rendering, exhibiting SOTA performance in various metrics compared to previous approaches.




\section{Our Approach}
Given \(N\) unposed images $\mathbb{I}=\left\{\mathbf{I}_r \in\right.\left.\mathbb{R}^{H \times W \times 3} \mid r=1 \cdots N\right\}$ as references, our goal is to synthesize target (or called query) image $\mathbf{I}_t \in\mathbb{R}^{H \times W \times 3}$ from novel view with corresponding poses $\mathbb{T}=\left\{\mathbf{T}_{r\rightarrow t} \mid r=1 \cdots N\right\}$.
Our GGRt is designed to train a generalizable Gaussian splatting network in a self-supervised manner, without relying on any camera pose or depth acquired in advance.

\begin{figure}[tb]
  \centering
  \includegraphics[height=6cm]{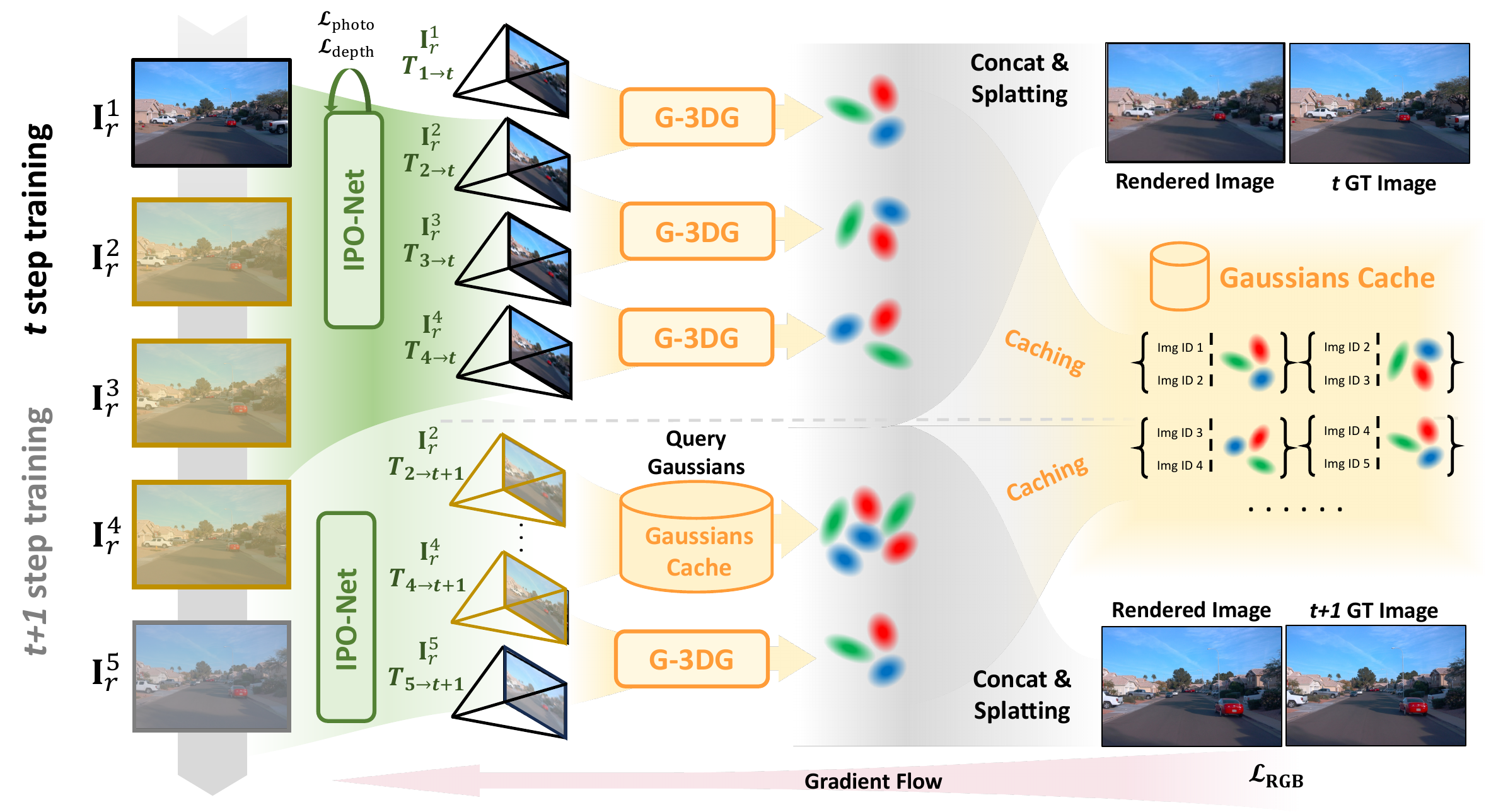}
  \caption{An overview of our method, demonstrated by using two continuous training steps given \(N\) selected nearby images. In the first training step, reference views are selected from nearby time \(r\in \mathcal{N} (t)\), then the IPO-Net estimates the relative poses between reference and target image {\color[HTML]{548235} 
 \(\mathbf{T}_{r\rightarrow t}\) }for 3D-Gaussian predictions. Then   \(\mathbf{I}^1_r \cdots \mathbf{I}^4_r\) forms three image pairs and is fed into the G-3DG model to predict Gaussians \(\mathbf{G}_1\cdots\mathbf{G}_3\) for novel view splatting and store them in Gaussians cache. In the second step, since {\color[HTML]{CCCC00}\(\mathbf{I}^2_r \cdots \mathbf{I}^4_r\)} are collaboratively used by the last step, we directly query their image ID in the Cache Gaussians and pick up corresponding Gaussian points \(\mathbf{G}_2,\mathbf{G}_3\) with newly predicted \(\mathbf{G}_4\) for novel view splatting. }
  \label{fig:framework}
\vspace{- 0.6cm}
\end{figure}

\subsection{The Joint Learning Framework}
\label{subsec:pipeline}
\subsubsection{Shared Image Encoder.}
We utilize the ResNet backbone pre-trained with DINO and augmented with feature pyramid networks (FPN) to extract both geometric and semantic cues from each reference view \(\mathbf{I}_r\) and the target view \(\mathbf{I}_t\). 
The extracted features are denoted as $\mathbf{F}_r$ and $\mathbf{F}_t$.

\subsubsection{Iterative Pose Optimization Network.}
 \label{subsec:pose} 
The goal of our pose estimation is to obtain the relative pose \(\mathbf{T}_{r \rightarrow t}\) between target and reference images. Hence the target image can be aggregated by the reference images. 
To this end, an intuitive solution is to build a cost function \(\mathcal{C}\) that enforces the feature-metric consistency across the target view and all nearby views (\textit{i.e.} minimize the re-projection error): 
\begin{equation}
\label{eq:reprojection}
\mathcal{C} =\frac{1}{|\mathcal{N} (t)|} \sum_{r\in \mathcal{N} (t)} \left\| \chi \left( \mathbf{T}_{r\rightarrow t} \mathbf{D}_{t} ,\mathbf{F}_{r} \right)  -\chi \left( \mathbf{u}_{t} ,\mathbf{F}_{t} \right)  \right\|,  
\end{equation}
where \(\chi\) denotes the interpolation function (e.g., bilinear interpolation), and $\mathbf{D}_{t}$ is the predicted depth of the target image.
Afterward, following the RAFT~\cite{teed2020raft} architecture, we adopt a Conv-GRU module to iterative update the camera pose $\mathbf{T}_{r\rightarrow t}$ and depth map $\mathbf{D}_t$. Specifically, at the iteration step $k=0$, given camera poses \(\mathbf{T}^{k=0}_{r\rightarrow t}\) and depth map \(\mathbf{D}^{k=0}_{t}\), we compute an initial cost map \(\mathcal{C}^{k=0}\) using Eq. \ref{eq:reprojection}. Then, the Conv-GRU module predicts the relative camera pose difference \(\Delta\mathbf{T}_{r\rightarrow t}\) and depth difference  \(\Delta\mathbf{D}_{t}\) to update the camera poses and the depth map for a predefined maximal number of iterations $K$, such that:

\begin{equation}
\mathbf{T}_{r\rightarrow t}^{(k)}=\mathbf{T}_{r\rightarrow t}^{(k-1)}+\Delta \mathbf{T}_{r\rightarrow t}, \quad \mathbf{D}_{t}^{(k)}=\mathbf{D}_t^{(k-1)}+\Delta \mathbf{D}_t.
\end{equation}
Then we transfer our pre-estimated relative poses  \(\mathbf{T}_{r\rightarrow t}\) between reference views and target view to relative poses \(\mathbf{T}_{r\rightarrow r+1}\) between nearby references view for our Gaussisans prediction. We denote the network as IPO-Net in the following context.

\subsubsection{Generalizable 3D-Gaussians.}
\label{subsec:gaussian}
Unlike previous generalizable methods that rely on implicit neural rendering and require ray aggregation for each target view, our approach is based on 3D-GS, which employs an explicit representation. As a result, we can generate Gaussian points from reference views and combine them to render a larger scene.
To accomplish this, we organize the image set $\mathbb{I}$ into several image pairs as well as their relative poses \(\{(\mathbf{I}^1_r,\mathbf{I}^2_r,\mathbf{T}_{1\rightarrow 2}),...,(\mathbf{I}^{N-1}_r,\mathbf{I}^{N}_r),\mathbf{T}_{N-1\rightarrow N}\}\)  and perform pixel-aligned Gaussian prediction $\mathbf{G}_i=\left\{\mathbf{g}_k=\right.\left.\left(\boldsymbol{\mu}_k, \boldsymbol{\Sigma}_k, \boldsymbol{\alpha}_k, \mathbf{S}_k\right)\right\}_k^K$ for each image pair  $(\mathbf{I}_r^i,\mathbf{I}^{i+1}_r)$.

In particular, given an image pair $(\mathbf{I}^i_r, \mathbf{I}^{i+1}_r)$, we design a module dubbed Generalizable 3D-Gaussians(G-3DG) to predict Gaussian points, which consists of three parts: 1) Epipolar Sampler, 2) Cross-Attention module, and 3) Local Self-Attention module. 
Let \(\mathbf{u}^i\) be the pixel coordinate from \(\mathbf{F}^i_r\) and \(\ell\) be the epipolar line induced by its ray in \(\mathbf{F}^{i+1}_r\)  . First, along  \(\ell \) , we sampled the features \(\mathbf{F}^{i+1}_r[\mathbf{u}_\ell^{i+1}]\) and the annotated points in it with the corresponding depths  \(\mathbf{D}_\ell^{i+1} \):
\begin{equation}
\label{eq:epipolar}
S(\mathbf{F}^{i+1}_{r})=\mathbf{F}^{i+1}_r[\mathbf{u}_\ell^{i+1}] \oplus \mathcal{PE}\left(\mathbf{D}_\ell^{i+1}\right),
\end{equation}
where \(\oplus\) and \(\mathcal{PE}(\cdot)\) indicates concatenation and positional encoding. Subsequently, we employ the Cross-Attention module $\mathcal{CA}(\cdot)$ to determine per-pixel correspondence. The feature \(\hat{\mathbf{F}_i}\)  incorporates a weighted sum of the depth positional encoding, with the expectation that the highest weight corresponds to the correct correspondence.
\begin{equation}
\label{eq:cross-attention}
    \hat{\mathbf{F}^i_r} = \mathbf{F}^i_r+\mathcal{CA}\left(q=\mathbf{F}^i_r, k=S(\mathbf{F}^{i+1}_{r}), v=S(\mathbf{F}^{i+1}_{r}) \right).
\end{equation}
Furthermore, we use a self-attention module to ensure our network propagates scaled depth estimation to part of the \(i\)-th image feature maps that may not have any epipolar correspondences in the  $(i+1)$-th image. For high-resolution novel view synthesis, we separate a high-resolution image into several small crops and utilize a Local Self-Attention module $\mathcal{LSA}(\cdot)$. This way, we can keep the training objective the same as the global self-attention trained on full images.
In detail, we split features \(\mathbf{F}^i_r\) into \(M \times M\) patches and conduct self-attention (\(\mathcal{SA}(\cdot)\)) for every patch. Then, we add positional encoding \(\mathcal{PE}(H,W)\) to retain the image-wise positional information, where \(H\) and \(W\) denote the height and width of the features \(\mathbf{F}^i\).
\begin{equation}
    \tilde{\mathbf{F}^i_r} = \mathcal{LSA}(\hat{\mathbf{F}^i_r}) = \hat{\mathbf{F}^i_r} + \begin{bmatrix}
\mathcal{SA}(\hat{\mathbf{F}}_{r,1,1}^i) & \mathcal{SA}(\hat{\mathbf{F}}_{r,1,2}^i) & \cdots & \mathcal{SA}(\hat{\mathbf{F}}_{r,1,m}^i) \\
\mathcal{SA}(\hat{\mathbf{F}}_{r,2,1}^i) & \mathcal{SA}(\hat{\mathbf{F}}_{r,2,2}^i) & \cdots & \mathcal{SA}(\hat{\mathbf{F}}_{r,2,m}^i) \\
\vdots & \vdots & \ddots & \vdots \\
\mathcal{SA}(\hat{\mathbf{F}}_{r,m,1}^i) & \mathcal{SA}(\hat{\mathbf{F}}_{r,m,2}^i) & \cdots & \mathcal{SA}(\hat{\mathbf{F}}_{r,m,m}^i)
\end{bmatrix} + \mathcal{PE}(H, W).
\end{equation}

Upon $ \tilde{\mathbf{F}^i_r} $, we predict its corresponding Gaussian points  $\mathbf{G}_i$ following the implementation of pixelSplat\cite{charatan2023pixelsplat}.
Our proposed training strategy naturally enables us to concatenate all the Gaussian points generated by image pairs for large scene generalization:
\begin{equation}
    \mathbb{G} = \left\{\mathbf{G}_1, \mathbf{G}_2,...,\mathbf{G}_{N-1}\right\}.
    \label{eq:concat}
\end{equation}
Moreover, our approach can achieve comparable performance while significantly reducing the required time for encoding reference images and therefore facilitating real-time rendering.

\subsubsection{Gaussians Cache Mechanism. }
As shown in Fig. \ref{fig:framework}, for two continuous training/inferencing iterations, many reference views \(\mathbf{I}^i_r\) of the current iteration are co-used by the last iteration. 
Re-inferencing them in the next iteration will be time-consuming and unnecessary since they share the same relative pose and features. 
Therefore, instead of re-predicting Gaussian points for all image pairs, we propose a dynamic store,  query, and release mechanism called Gaussians Cache. 
In \(t\)-th step,  it stores the predicted Gaussian points with corresponding image IDs \(\{i: \mathbf{G}_i\}\) in the cache. In \((t+1)\)-th step, it queries the Cache using image ID \(i\)  to restore the Gaussian points \(\mathbf{G}_i\). Furthermore, after querying all the IDs in the current iteration, the Cache releases the remaining unmatched Gaussians to optimize memory usage. This ensures that these Gaussians will not be utilized in the future, thereby reducing memory footprint without compromising performance.

\subsection{End-to-end Training with Deferred Optimization}

\subsubsection{Jointly Training Strategy.} 
\label{subsec:joint}
Learning scenes without camera pose is challenging due to the lack of 3D spatial priors to learn potential occlusions, varying lighting conditions, and the camera intrinsics encountered in unstructured environments.
To address this problem, as depicted in Figure \ref{fig:framework}(b), we activate the gradients of both the IPO-Net and G-3DG model, allowing for their simultaneous learning. By leveraging supervision from our rendered images, we can effectively optimize our IPO-Net.

Specifically, to strike a balance between pose estimation and training a generalized Gaussian network, we employ joint training of our GGRt model using the loss function $\mathcal{L}_{\text {joint }}$, which incorporates dynamic weight coefficient adjustment:
\begin{equation}
\mathcal{L}_{\text {joint }}=2^{\beta \cdot t}\left(\mathcal{L}_{\text {depth }}+\mathcal{L}_{\text {photo }}\right)+\left(1-2^{\beta \cdot t}\right) \mathcal{L}_{\text {rgb }},
\end{equation}
where $t$ denotes the training step. For IPO-Net optimization, we adopt photometric loss $\mathcal{L}_{\text {photo}}$~\cite{gu2021dro} and edge-aware smoothness $\mathcal{L}_{\text {depth}}$~\cite{godard2019digging} for our target view:

\begin{equation}
    \mathcal{L}_{\text {photo }}=\frac{1}{\left|\mathcal{N}_t\right|} \sum_{r \in \mathcal{N}_t}\left(\alpha \frac{1-\operatorname{ssim}\left(\mathbf{I}_t^{\prime}-\mathbf{I}_t\right)}{2}+(1-\alpha)\left\|\mathbf{I}_t^{\prime}-\mathbf{I}_t\right\|\right),
\end{equation}
\begin{equation}
\mathcal{L}_{\text {depth }}=\left|\partial_x \mathbf{D}\right| \exp ^{-\left|\partial_x \mathbf{I}\right|}+\left|\partial_y \mathbf{D}\right| \exp ^{-\left|\partial_y \mathbf{I}\right|},
\end{equation}
where \(\mathbf{I}_t^{\prime}\) denotes the warped image from the reference view \(r\) to target view \(t\). \(\partial_x\) and \(\partial_y\) are the image gradients. For our Gaussian model training, we simply apply MSE error between the rendered target image \(\hat{\mathbf{C}}\) and the ground truth target image \(\mathbf{C}\), where \(\mathbf{u}\) denotes pixel coordinate from image \(\mathbf{I}_t\):
\begin{equation}
\mathcal{L}_{\text {rgb }}=\sum_{\text{u} \in \mathbf{u}}\left\|\hat{\mathbf{C}}(\text{u})-\mathbf{C}(\text{u})\right\|_2^2.
\end{equation}

\subsubsection{Deferred Back-propagation for Generalizable Gaussians.} 
\begin{figure}[tb]
  \centering
  \includegraphics[width=0.8\linewidth]{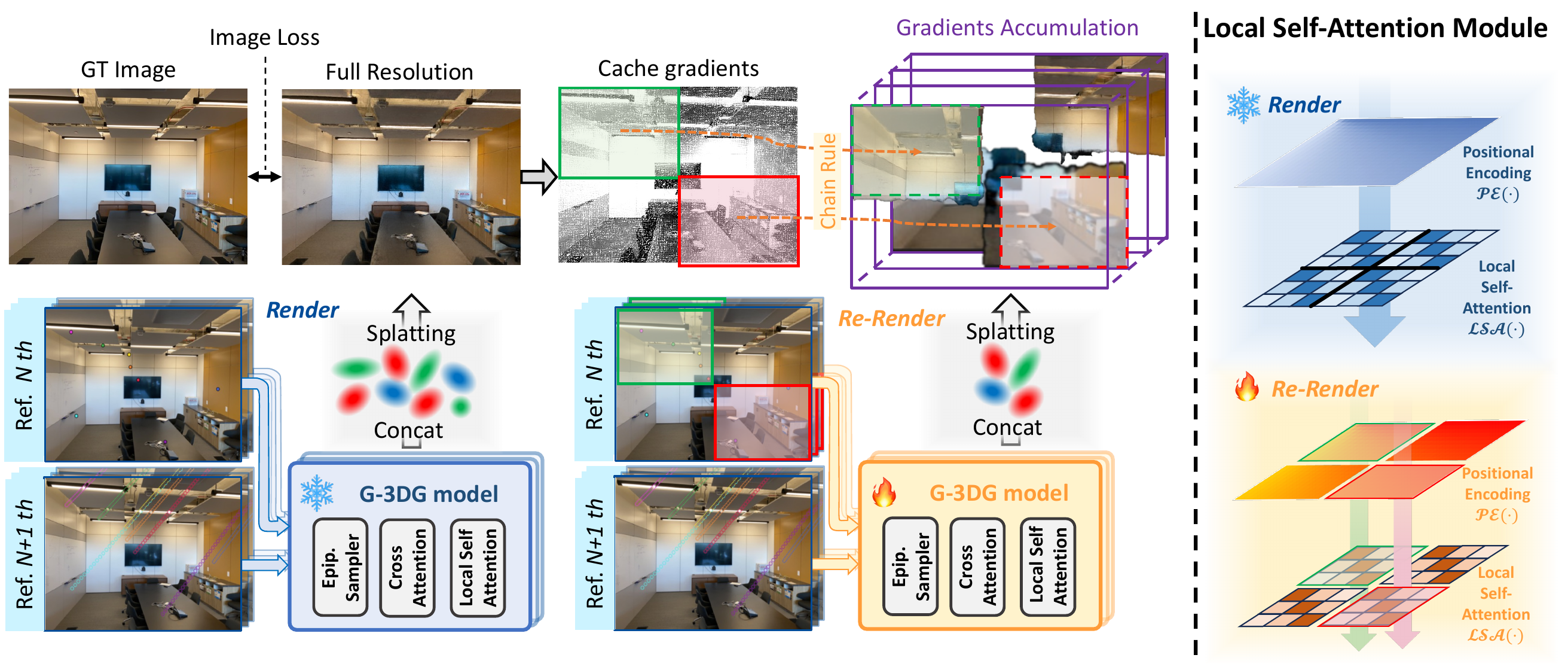}
  \caption{Illustration of deferred back-propagation pipeline of our G-3DG model (left column) and the procedure of our local self-attention module in deferred back-propagation (right column).  Details are shown in Sec. \ref{subsec:defer}. 
  }
  \label{fig:defer-pipeline}
\vspace{-0.6cm}
\end{figure}
\label{subsec:defer}
One of our goals is to render high-resolution images at a fast speed. 
However, with the growing image size, there will be insufficient memory to train a full high-resolution image due to the limited GPU memory. 
Inspired by~\cite{wang2023nerfart}, as shown in Fig. \ref{fig:defer-pipeline}, we specifically design \textit{deferred back-propagation} with G-3DG model that allows us to train GGRt in a high resolution under memory constraints.
Specifically, we generate a full-resolution image during the initial stage without utilizing auto-differentiation. Subsequently, we compute the image loss and its gradient on a per-pixel basis with regard to the rendered image.
During the second stage, we adopt a patch-wise approach to re-render pixels, enabling auto-differentiation. This allows us to back-propagate the cached gradients to the network parameters, facilitating their accumulation.

As mentioned earlier, the crucial aspect of ensuring successful training in the technique is maintaining consistency in our training objective across the two stages. Employing traditional self-attention mechanisms for both the entire image and patch images would result in an imbalance, as the features in the whole image can globally attend to information. In contrast, patch images can only focus on one patch at a time.
To solve this problem,  for every image pair $(\mathbf{I}^i_r, \mathbf{I}^{i+1}_r)$ in \(N\) reference images, we split the  \(\mathbf{I}^i_r\) into \(M\times M\) patches \([\mathbf{I}_{1,1}^i,\cdots,\mathbf{I}_{m,m}^i]\) , each patch follows the same shape with Local Self-Attention module elaborated above.  For each patch \(\mathbf{I}_{m,m}^i\), we rewrite the epipolar sampler of Eq. \ref{eq:epipolar} as: 
\begin{equation}
    S(\mathbf{F}^{i+1}_{r})=\mathbf{F}^{i+1}_r[\mathbf{u}_{\ell^{m,m}}^{i+1}] \oplus \gamma\left(d_{\ell^{m,m}}^{i+1}\right)
\end{equation}
where \(\ell^{m,m}\) denotes the epipolar lines induced by patch  \(\mathbf{I}^{m,m}_i\).  During the Cross-Attention module, we replace $q$ from \(\mathbf{F}^i_r\) to \(\mathbf{F}_{m,m}^i\) in Eq. \ref{eq:cross-attention}, which aims to aggregate global features from \(\mathbf{F}^{\mathbf{s}}_{i+1}\) for each pixel:
\begin{equation}
    \mathbf{F}_{m,m}^i = \mathbf{F}_{m,m}^i +  \mathcal{CA}\left(q=\mathbf{F}_{m,m}^i, k=S(\mathbf{F}^{i+1}_{r}), v=S(\mathbf{F}^{i+1}_{r}) \right)
\end{equation}
As for the self-attention module, firstly, we keep the positional encoding in global image-wise and crop them using the patch position to retain positional information between different patches. After that, the self-attention block $\mathcal{SA}(\cdot)$ is adopted for our patch feature $\mathbf{F}_{m,m}^i$:
\begin{equation}
    \mathbf{F}_{m,m}^i = \mathbf{F}_{m,m}^i + \mathcal{SA}(\mathbf{F}_{m,m}^i) + \text{Crop}(\mathcal{PE}(H, W),m,m)
\end{equation}
This design allows us to achieve consistent results between full image rendering and deferred rendering.

\subsection{Efficient Rendering} 
\label{sec:progressive}
By generating pixel-aligned Gaussian points in the reference views without being constrained by a specific target view, we can decompose the Gaussian generation process and the novel view splatting process. This decomposition enables us to render real-time at a high frame rate ($\ge$ 100 FPS). During practical implementation, we generate and cache all the Gaussian points in the scene (\textit{i.e.} Gaussians Cache). As a result, when presented with a query view within the scene, we can efficiently retrieve the corresponding Gaussians by utilizing the nearby reference images' IDs. This allows us to render a novel view quickly and accurately.


\section{Experiments}
\subsection{Implementation Details}
The experiments for novel view synthesis are conducted under two settings, i.e., the \emph{generalized} and \emph{finetuned} settings. Firstly, we train our model in several scenes and directly evaluate our model on test scenes (\textit{i.e.}, unseen scenes). Secondly, we finetune our generalized model on each unseen scene with a small number of steps and compare them with per-scene optimized NeRF methods.  

\subsubsection{Datasets.} Following~\cite{chen2023dbarf}, we train and evaluate our method on LLFF~\cite{llff}. To further demonstrate the capability of our model in general settings, we evaluate our performance in forward-facing outdoor datasets (\textit{i.e.} Waymo Open Dataset~\cite{waymo_open} and KITTI Dataset~\cite{kitti}).
 
\subsubsection{Parameters.}  We train our method end-to-end on datasets of multi-view unposed images using the Adam optimizer to minimize the overall loss  $\mathcal{L}_{\text {all }}$. The learning rates set for IPO-Net and G-3DG model are  \(5\times 10^{-4}\) and  \(2\times 10^{-5}\) respectively, decaying exponentially over the course of training. For the LLFF dataset, our training and rendering resolutions are set to \(378\times 504\). The number of reference views is 5 for the generalized setting and 10 for the finetuning setting. As for the Waymo dataset, the rendering resolution is set to \(640\times 960\) for both generalized and finetuning settings. The training resolutions are set to  \(196\times 288\) and  \(504\times 760\) respectively. For the KITTI dataset, we train and render with the same resolution \(176\times 612\). The number of reference views is set to 5 for both the generalized and finetuning settings.
We split our training images into 4 patches during deferred back-propagation.

\subsubsection{Metrics.} For render quality evaluation, Peak Signal-to-Noise Ratio (PSNR), Structural Similarity Index Measure (SSIM)~\cite{wang2004image}, and the Learned Perceptual Image Patch Similarity (LPIPS)~\cite{zhang2018unreasonable} are adopted.

\begin{table}[tb]
\caption{Quantitative performance of novel view synthesis on the Waymo~\cite{waymo_open} and LLFF~\cite{llff} datasets under generalized conditions. Entries in \textbf{bold} indicate the best performance in a pose-free context, while \highlight{highlighted} represents the best overall.}
\tiny   
\begin{tabular}{cc|cc|cc|cc|cc|cc|cc}
\toprule
\multicolumn{2}{c|}{\multirow{3}{*}{Scene}} & \multicolumn{4}{c|}{PSNR$\uparrow$}                              & \multicolumn{4}{c|}{SSIM$\uparrow$}                              & \multicolumn{4}{c}{LPIPS$\downarrow$}                             \\
\multicolumn{2}{c|}{}                        & \multicolumn{2}{c}{Pose-free \ding{56}} & \multicolumn{2}{c|}{Pose-free \ding{52}} & \multicolumn{2}{c}{Pose-free \ding{56}} & \multicolumn{2}{c|}{Pose-free \ding{52}} & \multicolumn{2}{c}{Pose-free \ding{56}} & \multicolumn{2}{c}{Pose-free \ding{52}} \\
\multicolumn{2}{c}{}                        & IBRNet    & PixelSplat    & DBARF       & Ours        & IBRNet    & PixelSplat    & DBARF       & Ours        & IBRNet    & PixelSplat    & DBARF       & Ours        \\ \midrule
\multirow{6}{*}{\begin{sideways}Waymo\end{sideways}}                                         & 003          & 31.40 & \cellcolor[HTML]{FFFFC7}31.45&       25.17& \textbf{31.20}                         & 0.917                        & \cellcolor[HTML]{FFFFC7}0.920 &       0.834&\textbf{0.912}                        & 0.127 & \cellcolor[HTML]{FFFFC7}0.101                        &       0.225& \textbf{0.137 }                        \\
                                & 19         & 30.06                         & 32.22 &       23.45& \cellcolor[HTML]{FFFFC7}\textbf{32.34}& 0.907                                                & 0.928 &       0.810& \cellcolor[HTML]{FFFFC7}\textbf{0.928} & 0.130                         & \cellcolor[HTML]{FFFFC7}0.082                        &       0.225& \textbf{0.110} \\
                                & 36         & 29.88                         & \cellcolor[HTML]{FFFFC7}33.58 &       22.59& \textbf{31.64} & 0.904                                                & \cellcolor[HTML]{FFFFC7}0.942 &       0.807& \textbf{0.919}& 0.150                         &\cellcolor[HTML]{FFFFC7}0.087                        &       0.254& \textbf{0.120} \\
                                & 69         & 30.24                         & 31.81 &       21.97& \cellcolor[HTML]{FFFFC7}\textbf{32.10} & 0.889                                                &\cellcolor[HTML]{FFFFC7}0.914 &       0.779& \textbf{0.908} & 0.175                         & \cellcolor[HTML]{FFFFC7}0.119                        &       0.307& \textbf{0.146} \\
                                & 81         & 31.30 & 31.02                         &       24.17&\cellcolor[HTML]{FFFFC7}\textbf{31.30} & 0.901                        & \cellcolor[HTML]{FFFFC7}0.904  &       0.785& \textbf{0.892}                         & 0.147 & {\color[HTML]{333333} \cellcolor[HTML]{FFFFC7}0.108} &       0.298& \textbf{0.159}                         \\ 
         & 126        & \cellcolor[HTML]{FFFFC7}36.00 & 33.68                         &       29.70& \textbf{34.15} & {\color[HTML]{333333} \cellcolor[HTML]{FFFFC7}0.940} & 0.931                         &       0.884& \textbf{0.937} & 0.106                         & 0.100      &       0.178&\cellcolor[HTML]{FFFFC7}\textbf{0.094}   \\ \midrule
         
\multirow{6}{*}{\begin{sideways}LLFF\end{sideways}}  &fern                     & 23.61 & 22.90& 23.12 & \cellcolor[HTML]{FFFFC7}\textbf{24.66} & 0.743 & 0.734 & 0.724                         & \cellcolor[HTML]{FFFFC7}\textbf{0.764} & 0.240 & \cellcolor[HTML]{FFFFC7}0.121 & 0.277 & \textbf{0.174} \\
& flower                   & 22.92 & 24.51 & 21.89                         & \cellcolor[HTML]{FFFFC7}\textbf{24.80} & \cellcolor[HTML]{FFFFC7}0.849 & 0.806 & 0.793                         & \textbf{0.795} & 0.123 & 0.106 & 0.176 &\cellcolor[HTML]{FFFFC7}\textbf{0.098} \\
& fortress                 &\cellcolor[HTML]{FFFFC7}29.05 & 26.36                         & 28.13 & \textbf{28.30} & \cellcolor[HTML]{FFFFC7}0.850 & 0.781                         & \textbf{0.820} & 0.817 & 0.087 & \cellcolor[HTML]{FFFFC7}0.080 & 0.126 & \textbf{0.109} \\
& horns                    & 24.96 & 23.86                         & 24.17 & \cellcolor[HTML]{FFFFC7}\textbf{26.34}& 0.831 & 0.836 & 0.799                         & \cellcolor[HTML]{FFFFC7}\textbf{0.872}& 0.144 & 0.127 & 0.194 & \cellcolor[HTML]{FFFFC7}\textbf{0.118 }\\
& leaves                   & 19.03 & 19.49 & 18.85                         &\cellcolor[HTML]{FFFFC7}\textbf{21.03} & \cellcolor[HTML]{FFFFC7}0.737 & 0.663 & 0.649                         & \textbf{0.729} & 0.289 & 0.158 & 0.313 & \cellcolor[HTML]{FFFFC7}\textbf{0.150} \\
& orchids                  & 18.52 & 17.65                         & 17.78 & \cellcolor[HTML]{FFFFC7}\textbf{19.00} & 0.573 & 0.533 & 0.506                         & \cellcolor[HTML]{FFFFC7}\textbf{0.589} & 0.259 & \cellcolor[HTML]{FFFFC7}0.191 & 0.352 & \textbf{0.219} \\
& room                     & 28.81 & 27.82 & 27.50                         & \cellcolor[HTML]{FFFFC7}\textbf{29.02} & 0.926 & 0.920 & 0.901                         & \cellcolor[HTML]{FFFFC7}\textbf{0.932} & 0.099 & \cellcolor[HTML]{FFFFC7}0.067 & 0.142 & \textbf{0.081}\\
& trex                     & 23.51 & 22.75 & 22.70                         & \cellcolor[HTML]{FFFFC7}\textbf{24.22} & \cellcolor[HTML]{FFFFC7}0.818 & 0.788 & 0.783                         & \textbf{0.804} & 0.160 & \cellcolor[HTML]{FFFFC7}0.107 & 0.207 & \textbf{0.166} \\ \bottomrule             
\end{tabular}
\label{tab:generalized}
\end{table}

\begin{table}[tb]
  \caption{Results for novel view synthesis on KITTI~\cite{kitti} dataset in generalized settings. Specifically, `w/o ft' denotes applying our method directly on KITTI using the weights trained on the Waymo dataset without finetuning, whereas `w/ ft' denotes the results from our model finetuned on KITTI in scene `04'. Resolution is \(176\times 612\). Time includes inferencing and rendering.}
  \label{tab:kitti}
\centering
\scriptsize

\begin{tabular}{c|cc|cccccc}
\toprule
\fontsize{8pt}{8pt}\selectfont
\multirow{2}{*}{Metric} & \multicolumn{2}{c|}{Pose-free \ding{56}} & \multicolumn{6}{c}{Pose-free \ding{52}}                                      \\
                         & IBRNet      & PixelSplat      & VideoAE & RUST  & FlowCAM & DBARF & Ours \textit{w/o} ft & Ours \textit{w/} ft \\ \midrule
PSNR $\uparrow$                     & 22.5        & \textbf{23.35}& 15.17   & 14.18 & 17.69   & 18.36 & 20.24          & \textbf{22.59}\\
LPIPS $\downarrow$                  & 0.44        & \textbf{0.129}& 0.462   & 0.654 & 0.405   & 0.425 & 0.388          & \textbf{0.327}\\ \midrule
 Time(s) $\downarrow$ & 0.850& \textbf{0.285}&  $\approx$ 2& $\approx$ 1& 4.170  & 0.850& \textbf{0.295}& \textbf{0.295}\\ \bottomrule
\end{tabular}
\vspace{-0.3cm}
\end{table}

\subsection{Benchmarking}
We first conduct experiments to compare our method with other methods, including both pose-required and pose-free, in light field dataset, LLFF~\cite{llff} and forward-facing autonomous driving dataset, Waymo Open dataset~\cite{waymo_open}.
As shown in Tab. \ref{tab:generalized}, our method achieves remarkable performance improvements compared to other approaches.  
Notably, our method comprehensively surpasses the best pose-free method DBARF~\cite{chen2023dbarf}, across all scenarios in both datasets. Specifically, in the `69' scene of the Waymo dataset, our method's PSNR exceeds that of DBARF by up to $10.3$ dB. 
Furthermore, compared to state-of-the-art pose-based methods like IBRNet~\cite{wang2021ibrnet} and pixelSplat~\cite{charatan2023pixelsplat}, our method also delivers highly competitive results. For instance, our method outperforms IBRNet and pixelSplat in most scenarios on the LLFF dataset. In the `horns' scene, our approach achieves a PSNR of $1.38$ dB and $2.48$ dB higher than IBRNet and pixelSplat, respectively.

\begin{figure}[tb]
  \centering
  \includegraphics[width=0.8\linewidth]{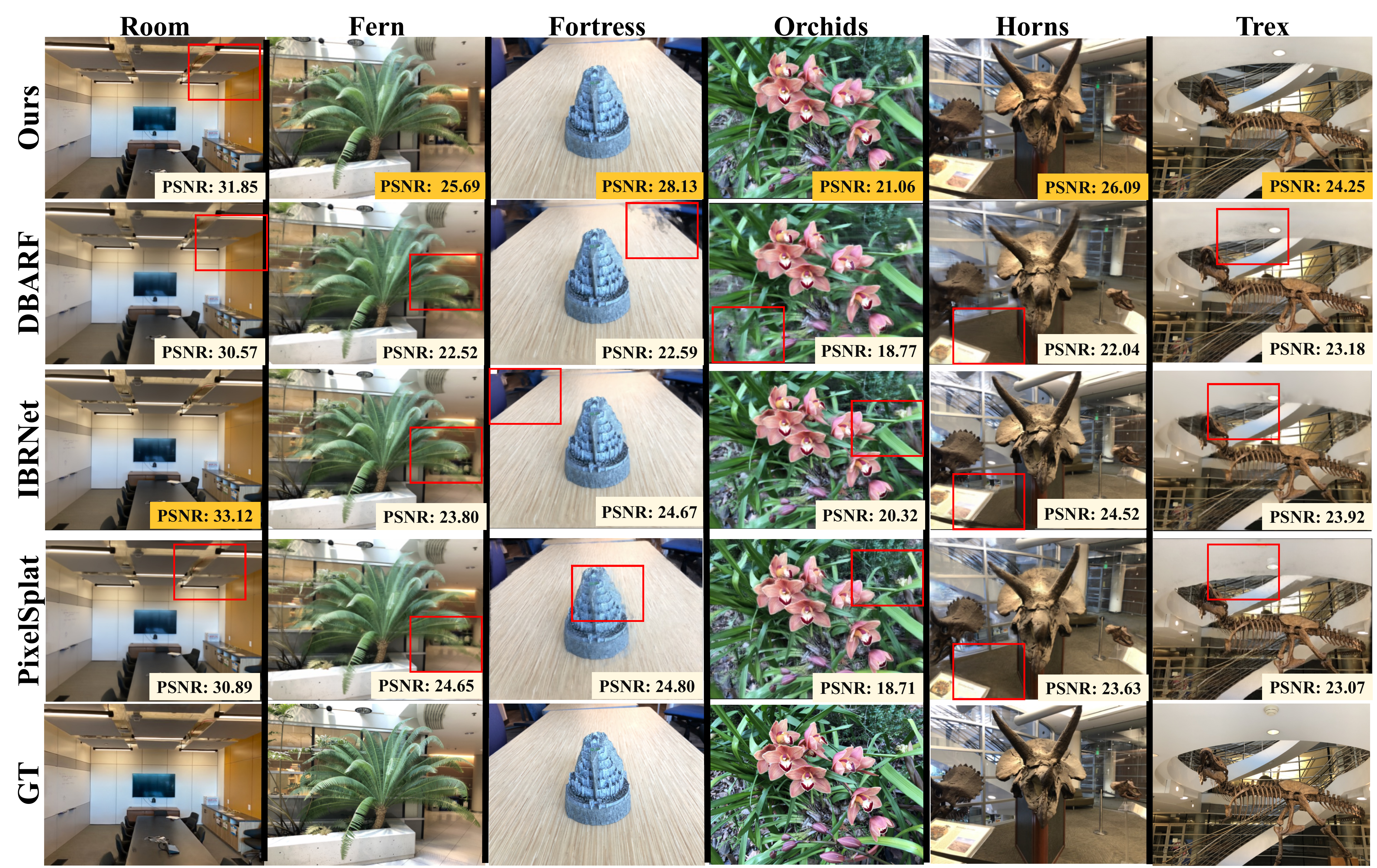}
  \caption{Novel view synthesis qualitative outcomes on the LLFF~\cite{llff} dataset under generalized settings, with significant regions highlighted by \textcolor{red}{red rectangles}.}
  \label{fig:llff_genearlized}
  \vspace{-0.3cm}
\end{figure}

\begin{figure}[tb]
  \centering
  \includegraphics[width=0.8\linewidth]{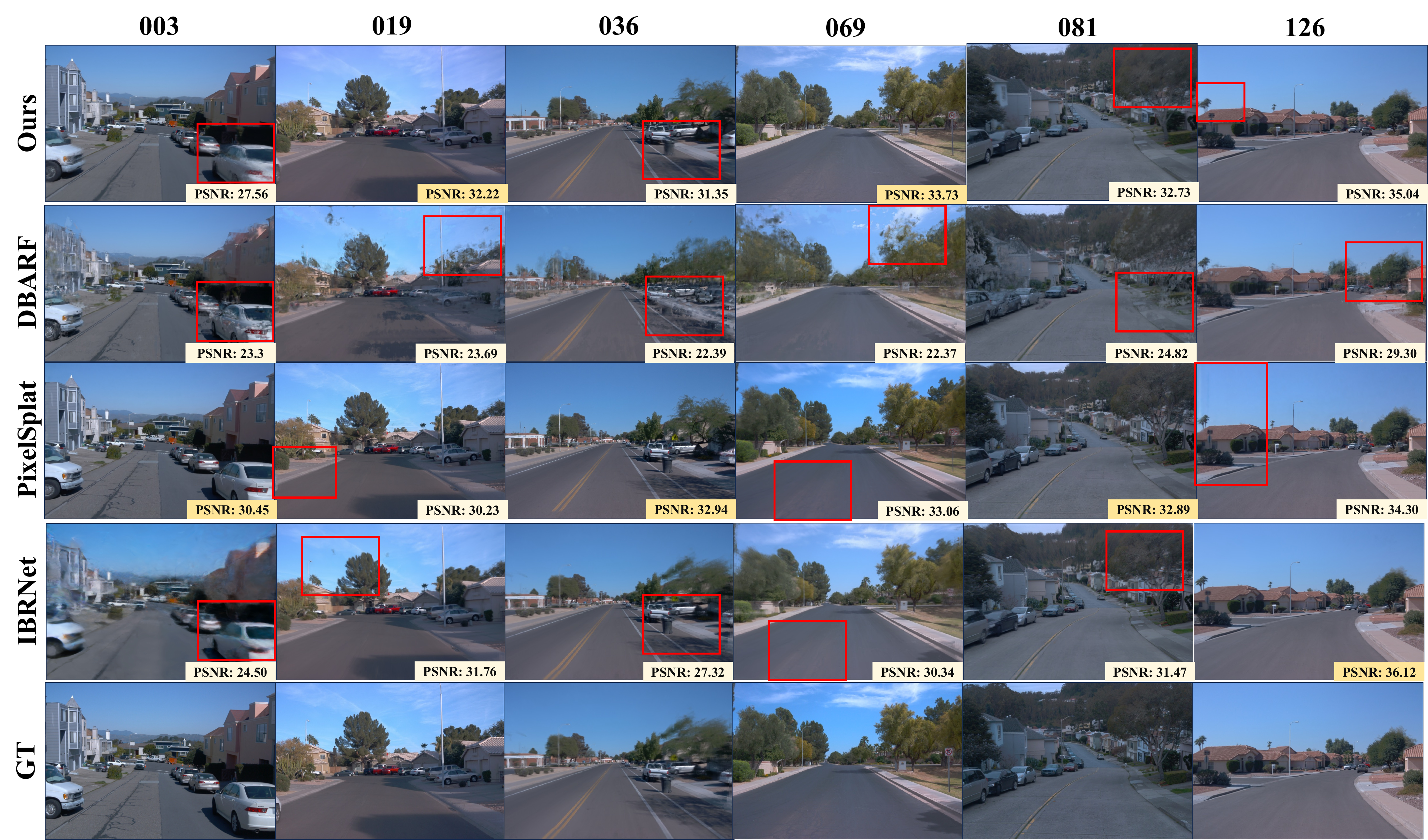}
  \caption{Qualitative results for novel view synthesis on Waymo~\cite{waymo_open} dataset with generalized settings. Areas of distinction are marked with \textcolor{red}{red rectangles}.}
  \label{fig:waymo_genearlized}
\end{figure}

We also conduct experiments on the KITTI~\cite{kitti} dataset to compare our method with other pose-free generalizable NeRF methods. 
As illustrated in Tab.~\ref{tab:kitti}, our approach outperforms the VideoAE~\cite{lai2021video}, RUST~\cite{sajjadi2023rust}, and FlowCAM~\cite{smith2023flowcam}, even without specifically training on the KITTI dataset. Notably, when we directly apply our Waymo-trained model to KITTI without any additional training, our method still surpasses those approaches that have been specifically trained on KITTI.
 Through fine-tuning our method on the KITTI dataset, we observe even more significant improvements. The PSNR values reach a remarkable value of $22.59$ dB, surpassing the state-of-the-art method FlowCAM by a substantial margin of $4.9$ dB.

\begin{figure}[tb]
  \centering
  \includegraphics[width=0.9\linewidth]{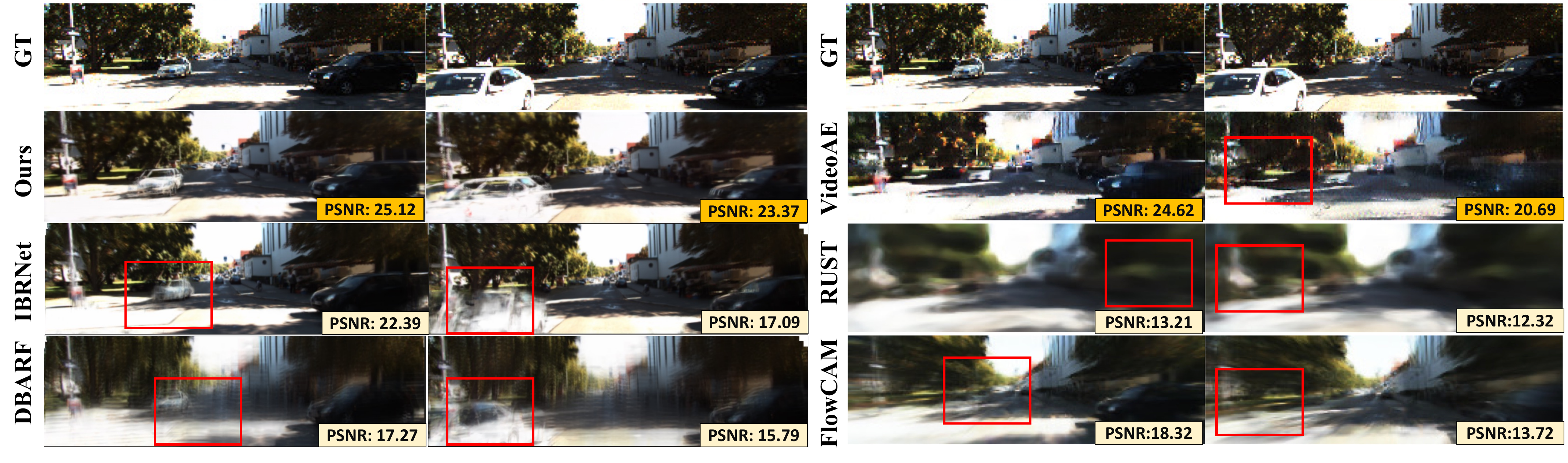}
  \caption{Results of novel view synthesis on the KITTI~\cite{kitti} dataset under generalized conditions, with \textcolor{red}{red rectangles} indicating areas of note. Note that ground truth (GT) is replicated for easier comparison.}
  \label{fig:kiti_genearlized}
\end{figure}

\subsubsection{Pose Accuracy Evaluation.} We assess the accuracy of our pose estimation on both quantitative and qualitative levels. Since we focus on estimating relative poses rather than absolute poses, we compare our results exclusively with DBARF. The comparison results are presented in Tab. \ref{tab:pose}, where we achieve better rotation and translation errors in most of the scenes. Notably, our rotation errors are significantly lower in the `Flower' and `Trex' scenes, consequently bringing significant reconstruction quality improvement by 2.91dB and 1.52dB in PSNR. 
\begin{figure}[t]
  \centering
  \includegraphics[width=10cm]{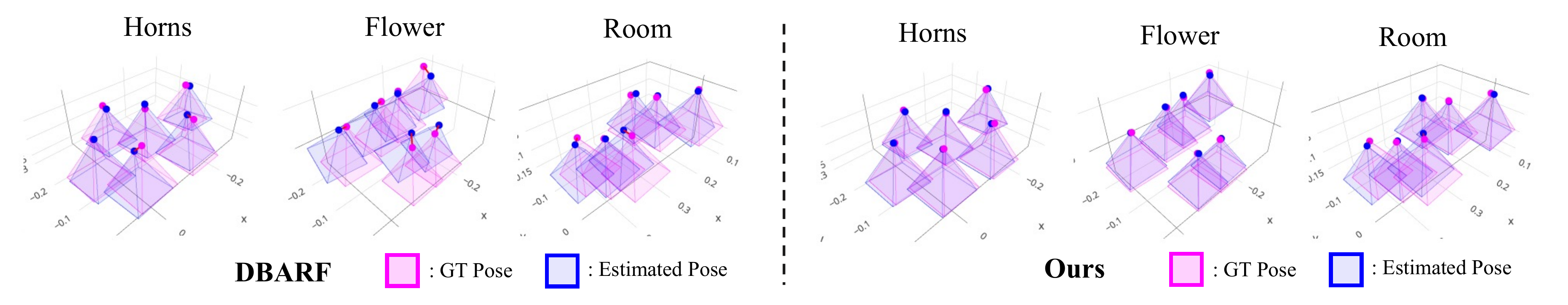}
  \caption{Qualitative results for pose optimization on three scenes of LLFF~\cite{llff} dataset.}
  \label{fig:pose_llff}
\end{figure}

\begin{table}[b]
\caption{Quantitative results of camera pose accuracy on LLFF~\cite{llff} dataset. Rotation denotes degree and translation is scaled by $10^2$.}
\centering
\fontsize{8pt}{8pt}\selectfont
\begin{tabular}{c c | c c c c c c c c} 
\toprule
Error & Method & Room & Trex & Flower & Fern & Fortress & Orchids & Leaves & Horns \\ \midrule
\multirow{2}{*}{Rotation$\downarrow$} & DBARF & 9.590 & 126.93 & 14.650 & 8.314 & 2.740 & 14.43 & \textbf{13.927} & 8.455 \\
 & Ours & \textbf{7.692} & \textbf{3.449} & \textbf{5.885} & \textbf{5.783} & \textbf{2.160} & \textbf{4.157} & 16.964 & \textbf{4.684} \\
 \midrule
\multirow{2}{*}{Translation$\downarrow$} & DBARF & 0.060 & 0.070& 0.010 & 0.020& 0.009 & 0.046 & 0.022 & 0.027 \\
 & Ours & \textbf{0.043} & \textbf{0.014} & \textbf{0.004} & \textbf{0.008} & \textbf{0.006} & \textbf{0.012} & \textbf{0.014} & \textbf{0.005} \\ \bottomrule

\end{tabular}
\label{tab:pose}
\end{table}

\subsection{Ablation Study}

\subsubsection{Effectiveness of Gaussians Cache.}~ In this ablation study, we compare our method's training and inferencing time consumption on a single RTX 3090 GPU and further evaluate its impact on the metric of PSNR. As shown in Tab. \ref{tab:ablation_progressive}, the proposed Gaussians Cache prevents the need for re-predicting Gaussians that were processed in the previous iteration, resulting in a \(2\times\) speed increase during training and \(8\times\) boost during inference when compared to the baseline without Gaussians cache. Additionally, the proposed caching technique has been shown to have no negative impact on performance, underscoring its efficacy.

\begin{table}[!t]
\caption{Ablation study evaluating the proposed Gaussians Cache mechanism. Training\&inference conducted on Waymo `019' dataset with a resolution of $228\times320$.}
\centering
\fontsize{8pt}{8pt}\selectfont
\small
\begin{tabular}{c|cc|c} 
\toprule
Gaussians Cache & Training Time$\downarrow$ & Inference Time$\downarrow$& PSNR$\uparrow$ \\ \midrule
\XSolidBrush & 4s/iter & 1.02s/iter & 32.34 \\
\Checkmark & 2s/iter & 0.13s/iter & 32.34 \\ \bottomrule
\end{tabular}
\label{tab:ablation_progressive}
\vspace{-0.3cm}
\end{table}

\subsubsection{Effectiveness of Deferred Back-Propagation.} We evaluate the performance influence of the proposed DBP technique. As shown in Tab. \ref{tab:PDB}, when using the regular back-propagation, a single RTX 3090 GPU can only process 2 reference views with the resolution of $384 \times 496$. If we want to use more reference views, we need to reduce the resolution of the observed images, which would bring undesired performance drops. However, when equipped with the proposed DBP technique, we can use 5 reference views with the resolution of $384 \times 496$ and gain $1.37-3.02$ dB in PSNR compared to the traditional approach. The performance gains become even more under the finetune setting. Notably, when continually increasing the number of reference views to 7, our approach reaches 31.5 dB in PSNR, setting a new state-of-the-art record.

\begin{figure}[!t]
  \centering
  \includegraphics[width=10cm]{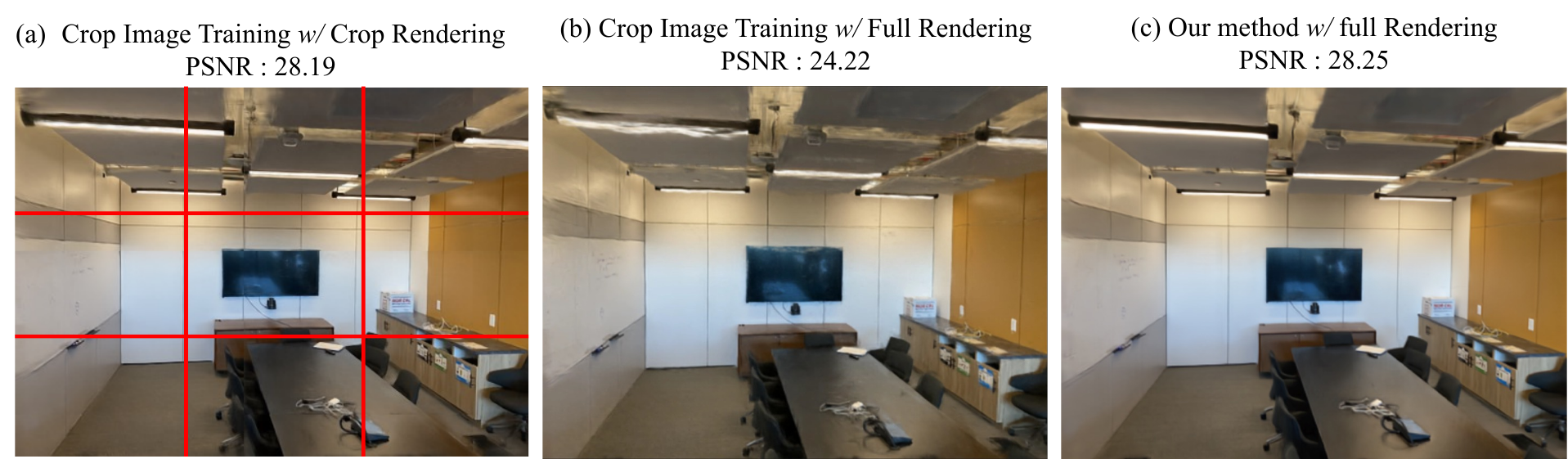}
  \caption{Visualization of our method and crop image training. (a) Individual image patches are rendered separately and stitched together. (b) The entire image is rendered using the weights trained with our crop image training approach. (c) The entire image is rendered using our deferred back-propagation technique.
  }
  \label{fig:crop}
  \vspace{-0.3cm}
\end{figure}

Apart from DBP, a simple idea to train a high-resolution model with limited hardware is to crop images from reference views randomly and only render the cropped areas.
We compared the training strategy with the Crop Image strategy and our DBP. After training the model with cropped images, it fails to capture the matching information of the entire image, resulting in poor rendering results when directly using a full image, which will. On the contrary, our proposed DBP works well under this circumstance as shown in Fig. \ref{fig:crop}.


\begin{figure}[tb]
  \centering
  \includegraphics[width=8cm]{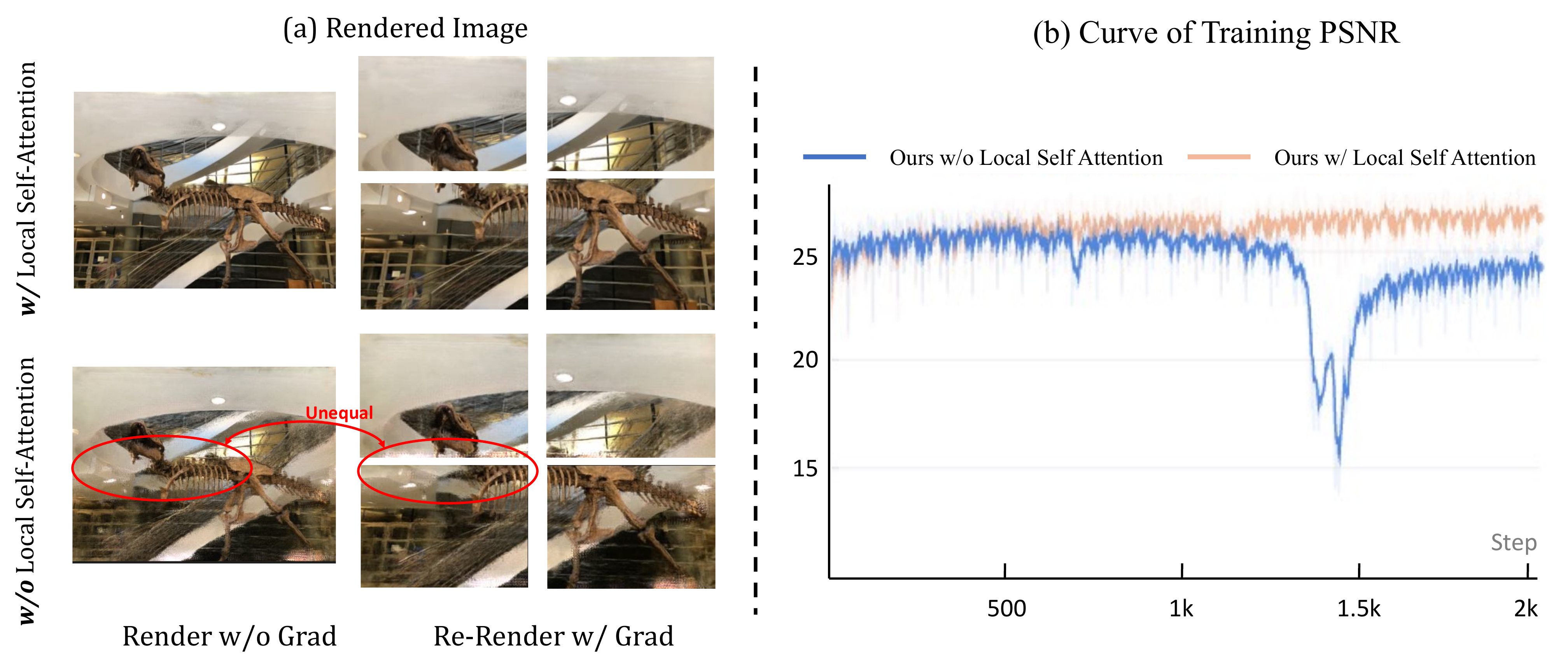}
  \caption{Ablation of our Local Self Attention for DBP. Left: the top and bottom row demonstrate the visualization of our DBP w/ or w/o local self-attention. 'Render w/o Grad' denotes the full image  ; Right: the PSNR over the course of training, a significant drop can be observed without our Local Self Attention.}
  \label{selfattention:fig}
\end{figure}

\begin{table}[tb]
\caption{Ablation study of our proposed deferred back-propagation technique on Scene `ROOM' from the LLFF dataset~\cite{llff}. `OOM' is an abbreviation for `out of memory'.}
\centering
\begin{tabular}{c|cc|cccc}
\toprule
                         &                             &                            & \multicolumn{2}{c}{PSNR $\uparrow$} & \multicolumn{2}{c}{Mem. (GB)} \\
\multirow{-2}{*}{Method} & \multirow{-2}{*}{Ref. View} & \multirow{-2}{*}{Resolution} & Gen.                     & Ft                     & Gen.           & Ft            \\ \midrule
Our \textit{w/o} Defer            & 2                           & $384\times 496$                         & 27.65                   & 27.77                  & 28.48         & 29.03         \\
Our \textit{w/o} Defer            & 3                           & $384\times 496$                         & -                       & -                      & OOM           & OOM           \\
Our \textit{w/o} Defer            & 5                           & $192\times 248$                         & 26.00                   & 28.51                  & 31.05         & 31.05         \\ \midrule
Our \textit{w/} Defer             & 5                           & $384\times 496$                         & 29.02                   & 29.85                  & 29.35         & 29.35         \\
Our \textit{w/} Defer             & 7                           & $384\times 496$                         & -                       & 31.50                  & -             & 34.61        \\ \bottomrule
\end{tabular}
\label{tab:PDB}
\vspace{-0.3cm}
\end{table}
\subsubsection{Effectiveness of Local Self-Attention.}
Here we conduct ablations of our proposed local self-attention. As shown in Fig. \ref{selfattention:fig}(a), without our local self-attention, the rendered images between the first and second steps of DBP are not consistent. Specifically, without the inclusion of Local Self-Attention, the PSNR performance experiences a drop after around 1,500 training steps, as illustrated in  Fig.~\ref{selfattention:fig} (b). To our knowledge, by utilizing Local Self-Attention, the results of both processes remain consistent. Otherwise, the loss calculated by the forward-only rendering process no longer serves as an appropriate guide for the gradient backpropagation of the DBP rendering process, resulting in a significant decrease in PSNR.

\section{Conclusion}
This paper introduces a novel method for generalizable novel view synthesis that eliminates the need for camera poses, enables high-resolution real-time rendering, and eliminates lengthy optimization. Our method contains jointly trained IPO-Net and G-3DG models, as well as the progressive Gaussian cache module, enabling robust relative pose estimation and fast scene reconstruction from image observations without prior poses. We incorporate a deferred back-propagation mechanism for high-resolution training and inference, overcoming GPU memory limitations. GGRt achieves impressive inferencing and real-time rendering speeds, outperforming existing pose-free techniques and approaching pose-based 3D-GS methods. Extensive experimentation on diverse datasets confirms its effectiveness.
%
%
\bibliographystyle{splncs04}
\bibliography{egbib}

\begin{thebibliography}{10}
\providecommand{\url}[1]{\texttt{#1}}
\providecommand{\urlprefix}{URL }
\providecommand{\doi}[1]{https://doi.org/#1}

\bibitem{bian2023nope}
Bian, W., Wang, Z., Li, K., Bian, J.W., Prisacariu, V.A.: Nope-nerf: Optimising neural radiance field with no pose prior. In: CVPR. pp. 4160--4169 (2023)

\bibitem{charatan2023pixelsplat}
Charatan, D., Li, S., Tagliasacchi, A., Sitzmann, V.: pixelsplat: 3d gaussian splats from image pairs for scalable generalizable 3d reconstruction. In: CVPR (2024)

\bibitem{chen2023dbarf}
Chen, Y., Lee, G.H.: Dbarf: Deep bundle-adjusting generalizable neural radiance fields. In: CVPR. pp. 24--34 (2023)

\bibitem{fu20233d}
Fu, Y., De~Mello, S., Li, X., Kulkarni, A., Kautz, J., Wang, X., Liu, S.: 3d reconstruction with generalizable neural fields using scene priors. In: ICLR (2024)

\bibitem{fu2023colmap}
Fu, Y., Liu, S., Kulkarni, A., Kautz, J., Efros, A.A., Wang, X.: Colmap-free 3d gaussian splatting. In: CVPR (2024)

\bibitem{kitti}
Geiger, A., Lenz, P., Urtasun, R.: Are we ready for autonomous driving? the kitti vision benchmark suite pp. 3354--3361 (2012)

\bibitem{godard2019digging}
Godard, C., Mac~Aodha, O., Firman, M., Brostow, G.J.: Digging into self-supervised monocular depth estimation. In: Proceedings of the IEEE/CVF international conference on computer vision. pp. 3828--3838 (2019)

\bibitem{gu2021dro}
Gu, X., Yuan, W., Dai, Z., Tang, C., Zhu, S., Tan, P.: {DRO}: Deep recurrent optimizer for video to depth. IEEE Robotics and Automation Letters  \textbf{8}(5),  2844--2851 (2023)

\bibitem{coponerf}
Hong, S., Jung, J., Shin, H., Yang, J., Kim, S., Luo, C.: Unifying correspondence, pose and nerf for pose-free novel view synthesis from stereo pairs. arXiv preprint arXiv:2312.07246  (2023)

\bibitem{hong2023lrm}
Hong, Y., Zhang, K., Gu, J., Bi, S., Zhou, Y., Liu, D., Liu, F., Sunkavalli, K., Bui, T., Tan, H.: Lrm: Large reconstruction model for single image to 3d. In: ICLR (2024)

\bibitem{kerbl20233d}
Kerbl, B., Kopanas, G., Leimk{\"u}hler, T., Drettakis, G.: 3d gaussian splatting for real-time radiance field rendering. ACM TOG  \textbf{42}(4) (2023)

\bibitem{lai2021video}
Lai, Z., Liu, S., Efros, A.A., Wang, X.: Video autoencoder: self-supervised disentanglement of static 3d structure and motion. In: ICCV. pp. 9730--9740 (2021)

\bibitem{li2023instant3d}
Li, J., Tan, H., Zhang, K., Xu, Z., Luan, F., Xu, Y., Hong, Y., Sunkavalli, K., Shakhnarovich, G., Bi, S.: Instant3d: Fast text-to-3d with sparse-view generation and large reconstruction model. In: ICLR (2024)

\bibitem{lin2021barf}
Lin, C.H., Ma, W.C., Torralba, A., Lucey, S.: Barf: Bundle-adjusting neural radiance fields. In: ICCV. pp. 5741--5751 (2021)

\bibitem{liu2022neural}
Liu, Y., Peng, S., Liu, L., Wang, Q., Wang, P., Theobalt, C., Zhou, X., Wang, W.: Neural rays for occlusion-aware image-based rendering. In: CVPR. pp. 7824--7833 (2022)

\bibitem{meuleman2023progressively}
Meuleman, A., Liu, Y.L., Gao, C., Huang, J.B., Kim, C., Kim, M.H., Kopf, J.: Progressively optimized local radiance fields for robust view synthesis. In: CVPR. pp. 16539--16548 (2023)

\bibitem{llff}
Mildenhall, B., Srinivasan, P.P., Ortiz-Cayon, R., Kalantari, N.K., Ramamoorthi, R., Ng, R., Kar, A.: Local light field fusion: Practical view synthesis with prescriptive sampling guidelines. ACM TOG  \textbf{38}(4),  1--14 (2019)

\bibitem{mildenhall2021nerf}
Mildenhall, B., Srinivasan, P.P., Tancik, M., Barron, J.T., Ramamoorthi, R., Ng, R.: Nerf: Representing scenes as neural radiance fields for view synthesis. Communications of the ACM  \textbf{65}(1),  99--106 (2021)

\bibitem{srt22}
Sajjadi, M.S.M., Meyer, H., Pot, E., Bergmann, U., Greff, K., Radwan, N., Vora, S., Lucic, M., Duckworth, D., Dosovitskiy, A., Uszkoreit, J., Funkhouser, T., Tagliasacchi, A.: {Scene Representation Transformer: Geometry-Free Novel View Synthesis Through Set-Latent Scene Representations}. In: CVPR (2022)

\bibitem{sajjadi2023rust}
Sajjadi, M.S., Mahendran, A., Kipf, T., Pot, E., Duckworth, D., Lu{\v{c}}i{\'c}, M., Greff, K.: Rust: Latent neural scene representations from unposed imagery. In: CVPR. pp. 17297--17306 (2023)

\bibitem{sitzmann2021light}
Sitzmann, V., Rezchikov, S., Freeman, B., Tenenbaum, J., Durand, F.: Light field networks: Neural scene representations with single-evaluation rendering. Advances in Neural Information Processing Systems  \textbf{34},  19313--19325 (2021)

\bibitem{smith2023flowcam}
Smith, C., Du, Y., Tewari, A., Sitzmann, V.: Flowcam: Training generalizable 3d radiance fields without camera poses via pixel-aligned scene flow. In: NeurIPS (2023)

\bibitem{suhail2022light}
Suhail, M., Esteves, C., Sigal, L., Makadia, A.: Light field neural rendering. In: CVPR. pp. 8269--8279 (2022)

\bibitem{waymo_open}
Sun, P., Kretzschmar, H., Dotiwalla, X., Chouard, A., Patnaik, V., Tsui, P., Guo, J., Zhou, Y., Chai, Y., Caine, B., Vasudevan, V., Han, W., Ngiam, J., Zhao, H., Timofeev, A., Ettinger, S., Krivokon, M., Gao, A., Joshi, A., Zhang, Y., Shlens, J., Chen, Z., Anguelov, D.: Scalability in perception for autonomous driving: Waymo open dataset. In: CVPR (June 2020)

\bibitem{teed2020raft}
Teed, Z., Deng, J.: Raft: Recurrent all-pairs field transforms for optical flow. In: ECCV. pp. 402--419. Springer (2020)

\bibitem{tian2023mononerf}
Tian, F., Du, S., Duan, Y.: Mononerf: Learning a generalizable dynamic radiance field from monocular videos. In: ICCV. pp. 17903--17913 (2023)

\bibitem{wang2022attention}
Wang, P., Chen, X., Chen, T., Venugopalan, S., Wang, Z., et~al.: Is attention all nerf needs? In: ICLR (2023)

\bibitem{wang2023pf}
Wang, P., Tan, H., Bi, S., Xu, Y., Luan, F., Sunkavalli, K., Wang, W., Xu, Z., Zhang, K.: Pf-lrm: Pose-free large reconstruction model for joint pose and shape prediction. In: ICLR (2024)

\bibitem{wang2021ibrnet}
Wang, Q., Wang, Z., Genova, K., Srinivasan, P.P., Zhou, H., Barron, J.T., Martin-Brualla, R., Snavely, N., Funkhouser, T.: {IBRNet}: Learning multi-view image-based rendering. In: CVPR. pp. 4690--4699 (2021)

\bibitem{wang2004image}
Wang, Z., Bovik, A.C., Sheikh, H.R., Simoncelli, E.P.: Image quality assessment: from error visibility to structural similarity. IEEE TIP  \textbf{13}(4),  600--612 (2004)

\bibitem{wang2021nerf}
Wang, Z., Wu, S., Xie, W., Chen, M., Prisacariu, V.A.: Nerf--: Neural radiance fields without known camera parameters. arXiv preprint arXiv:2102.07064  (2021)

\bibitem{yao2018mvsnet}
Yao, Y., Luo, Z., Li, S., Fang, T., Quan, L.: Mvsnet: Depth inference for unstructured multi-view stereo. In: ECCV. pp. 767--783 (2018)

\bibitem{yen2021inerf}
Yen-Chen, L., Florence, P., Barron, J.T., Rodriguez, A., Isola, P., Lin, T.Y.: {iNeRF}: Inverting neural radiance fields for pose estimation. In: IROS. pp. 1323--1330. IEEE (2021)

\bibitem{yu2021pixelnerf}
Yu, A., Ye, V., Tancik, M., Kanazawa, A.: {pixelNeRF}: Neural radiance fields from one or few images. In: CVPR. pp. 4578--4587 (2021)

\bibitem{wang2023nerfart}
Zhang, K., Kolkin, N., Bi, S., Luan, F., Xu, Z., Shechtman, E., Snavely, N.: Arf: Artistic radiance fields. In: ECCV. pp. 717--733. Springer (2022)

\bibitem{zhang2018unreasonable}
Zhang, R., Isola, P., Efros, A.A., Shechtman, E., Wang, O.: The unreasonable effectiveness of deep features as a perceptual metric. In: CVPR. pp. 586--595 (2018)

\end{thebibliography}
\end{document}


\title{GGRt: Towards Pose-free Generalizable 3D Gaussian Splatting in Real-time\\ Supplementary Material} 

\titlerunning{Abbreviated paper title}

\author{First Author\inst{1}\orcidlink{0000-1111-2222-3333} \and
Second Author\inst{2,3}\orcidlink{1111-2222-3333-4444} \and
Third Author\inst{3}\orcidlink{2222--3333-4444-5555}}

\authorrunning{F.~Author et al.}

\institute{Princeton University, Princeton NJ 08544, USA \and
Springer Heidelberg, Tiergartenstr.~17, 69121 Heidelberg, Germany
\email{lncs@springer.com}\\
\url{http://www.springer.com/gp/computer-science/lncs} \and
ABC Institute, Rupert-Karls-University Heidelberg, Heidelberg, Germany\\
\email{\{abc,lncs\}@uni-heidelberg.de}}

\maketitle

This supplementary material provides additional results on per-scene optimization (Section~\ref{sec:pso}), qualitative comparisons of depth results with other baselines (Section~\ref{sec:depth_vis}), and showcase more results of our proposed method.

\section{Per-scene Optimization Results}
\label{sec:pso}
This section assesses the effectiveness of our method through per-scene optimization evaluations conducted on the LLFF and Waymo Open datasets, respectively.

\subsection{LLFF Dataset Evaluation}
The per-scene optimization on the LLFF dataset involves iterating over 2,000 steps, as outlined in Table \ref{tab:finetune}. Our findings indicate that our method not only achieves comparable performance to generalized methods like IBRNet~\cite{wang2021ibrnet} and pixelSplat~\cite{charatan2023pixelsplat}, but also outperforms per-scene optimization approaches such as BARF~\cite{lin2021barf}, GARF~\cite{chng2022garf}, and DBARF~\cite{chen2023dbarf}. Through our evaluations, we observe that our method consistently demonstrates superior performance.
\begin{table}[tb]
    \caption{Quantitative results for novel view synthesis on LLFF \cite{llff} dataset. Entries in \textbf{bold} indicate the best performance in a pose-free context, while \highlight{highlighted} indicates the best overall performance.}
    \small
    \centering
\begin{tabular}{cl|cc|cccc}
\toprule
\multirow{2}{*}{Metric} & \multirow{2}{*}{Scenes} & \multicolumn{2}{c|}{Pose-free \ding{56}}            & \multicolumn{4}{c}{Pose-free \ding{52}}                                                                                                                                      \\
                        &                         & IBRNet                        & pixelSplat & BARF                                   & GARF                                   & DBARF                                  & Ours                                    \\ \midrule
\multirow{8}{*}{PSNR \(\uparrow\)}   & fern                    & 25.56                         & 23.91      & 23.79                                  & 24.51                                  & \cellcolor[HTML]{FFFFC7}\textbf{25.97} & 25.41                                   \\
                        & flower                  & 23.94                         & 25.02      & 23.37                                  & \cellcolor[HTML]{FFFFC7}\textbf{26.40} & 23.95                                  & 25.79                                   \\
                        & fortress                & 31.18                         & 26.97      & 29.08                                  & 29.09                                  & \cellcolor[HTML]{FFFFC7}\textbf{31.43} & 30.93                                   \\
                        & horns                   & 28.46                         & 25.04      & 22.78                                  & 23.03                                  & 27.51                                  & \cellcolor[HTML]{FFFFC7}\textbf{28.54}  \\
                        & leaves                  & 21.28                         & 21.16      & 18.78                                  & 19.72                                  & 20.32                                  & \cellcolor[HTML]{FFFFC7}\textbf{22.38}  \\
                        & orchids                 & \cellcolor[HTML]{FFFFC7}20.83 & 18.73      & 19.45                                  & 19.37                                  & 20.26                                  & \textbf{20.49}                          \\
                        & room                    & 31.05                         & 27.77      & \cellcolor[HTML]{FFFFC7}\textbf{31.95} & 31.90                                  & 31.09                                  & 31.50                                   \\
                        & trex                    & \cellcolor[HTML]{FFFFC7}26.52 & 24.22      & 22.55                                  & 22.86                                  & 22.82                                  & \textbf{25.75}                          \\ \midrule
\multirow{8}{*}{SSIM \(\uparrow\)}   & fern                    & \cellcolor[HTML]{FFFFC7}0.825 & 0.755      & 0.710                                  & 0.740                                  & 0.84                                   & \textbf{0.804}                          \\
                        & flower                  & 0.895                         & 0.817      & 0.698                                  & 0.790                                  & \cellcolor[HTML]{FFFFC7}\textbf{0.895} & 0.831                                   \\
                        & fortress                & 0.918                         & 0.833      & 0.823                                  & 0.820                                  & \cellcolor[HTML]{FFFFC7}\textbf{0.918} & 0.908                                   \\
                        & horns                   & 0.913                         & 0.857      & 0.727                                  & 0.730                                  & 0.903                                  & \cellcolor[HTML]{FFFFC7}\textbf{0.917 } \\
                        & leaves                  & \cellcolor[HTML]{FFFFC7}0.807 & 0.745      & 0.537                                  & 0.610                                  & 0.758                                  & \textbf{0.796}                          \\
                        & orchids                 & \cellcolor[HTML]{FFFFC7}0.722 & 0.583      & 0.574                                  & 0.570                                  & \textbf{0.693}                         & 0.665                                   \\
                        & room                    & 0.950                         & 0.915      & 0.940                                  & 0.940                                  & 0.947                                  & \cellcolor[HTML]{FFFFC7}\textbf{0.952}  \\
                        & trex                    & \cellcolor[HTML]{FFFFC7}0.905 & 0.834      & 0.767                                  & 0.800                                  & 0.848                                  & \textbf{0.871}                          \\ \midrule
\multirow{8}{*}{LPIPS \(\downarrow\)}  & fern                    & 0.139                         & 0.123      & 0.311                                  & 0.290                                  & \cellcolor[HTML]{FFFFC7}\textbf{0.120}                                  & 0.176                                   \\
                        & flower                  & 0.074                         & 0.096      & 0.211                                  & 0.110                                  & 0.084                                  & \cellcolor[HTML]{FFFFC7}\textbf{0.073}                                   \\
                        & fortress                & 0.046                         & 0.082      & 0.132                                  & 0.150                                  & \cellcolor[HTML]{FFFFC7}\textbf{0.046}                                  & 0.051                                   \\
                        & horns                   & 0.070                         & 0.106      & 0.298                                  & 0.290                                  & 0.076                                  & \cellcolor[HTML]{FFFFC7}\textbf{0.061}                                   \\
                        & leaves                  & 0.137                         & 0.176      & 0.353                                  & 0.270                                  & 0.156                                  & \cellcolor[HTML]{FFFFC7}\textbf{0.128}                                   \\
                        & orchids                 & \cellcolor[HTML]{FFFFC7}0.142                         & 0.236      & 0.291                                  & 0.260                                  & \textbf{0.151}                                  & 0.203                                   \\
                        & room                    & 0.060                         & 0.110      & 0.099                                  & 0.130                                  & 0.063                                  & \cellcolor[HTML]{FFFFC7}\textbf{0.051}                                   \\
                        & trex                    & \cellcolor[HTML]{FFFFC7}0.074                         & 0.1117     & 0.206                                  & 0.190                                  & 0.120                                  & \textbf{0.106}                                   \\ \bottomrule
\end{tabular}
    \label{tab:finetune}
\end{table}
Moreover, our method exhibits remarkable superiority over other pose-free approaches, while also delivering competitive results compared to pose-based methods. Specifically, in scenes such as `horns' and `leaves', our method surpasses all other evaluated approaches, both in generalized and pose-free scenarios. To provide visual evidence of these findings, qualitative results are presented in Figure \ref{fig:llff_finetune}. These results highlight the performance and robustness of our method across various challenging scenes.
\begin{figure}[H]
  \centering
  \includegraphics[width=12cm]{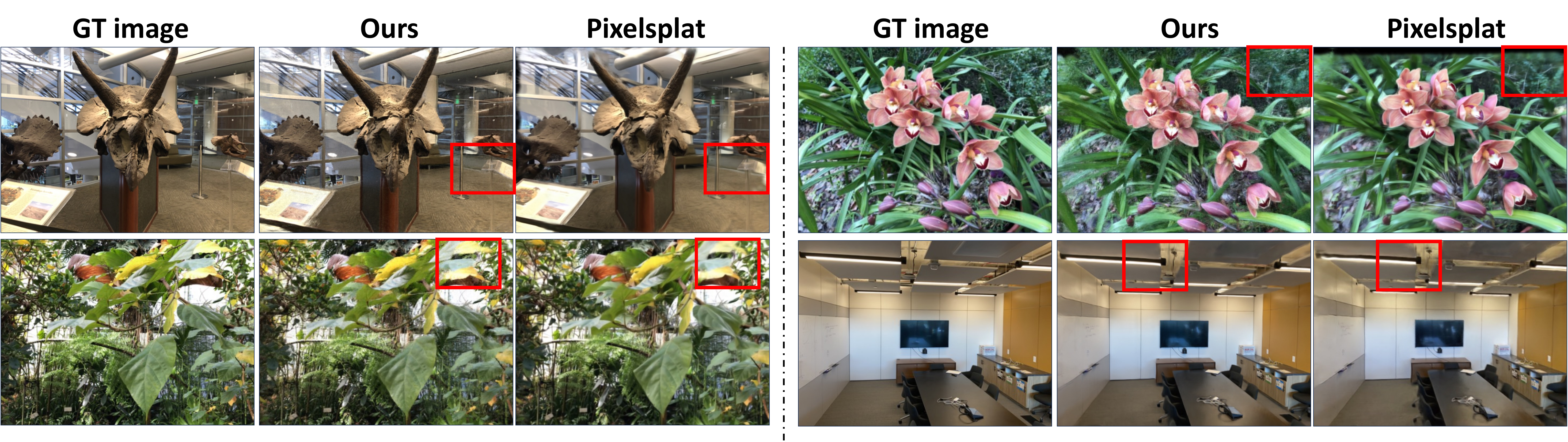}
  \caption{Qualitative comparison between pixelSplat~\cite{charatan2023pixelsplat} and our method on LLFF~\cite{llff} dataset under the fine-tuning setting. Areas of distinction are highlighted by \textcolor{red}{rectangles}.
  }
  \label{fig:pixel_our}
\end{figure}

Significantly, our method outperforms pixelSplat~\cite{charatan2023pixelsplat} across all scenes, including improvements of 1.5dB in `ferns' scene, 3.73dB in `room' scene, and 3.5dB in `horns' scene. This superiority can be attributed to our method's ability to aggregate multiple images ($7$ compared to $2$ in pixelSplat), enabling enhanced performance during per-scene optimization. To visually demonstrate the impact of these quantitative results, we provide a visualization comparison between our method and pixelSplat in Figure \ref{fig:pixel_our}. These results serve to reinforce the findings presented in Table \ref{tab:finetune}, showcasing the clear advantage of our method over pixelSplat in terms of quality and accuracy.

\begin{figure}[H]
  \centering
  \includegraphics[width=12cm]{figures/ft_llff.pdf}
  \caption{Qualitative results for novel view synthesis on LLFF~\cite{llff} dataset with finetuning settings.
  }
  \label{fig:llff_finetune}
\end{figure}

\subsection{Waymo Dataset Evaluation}
Additionally, we perform similar experiments on the Waymo \cite{waymo_open} dataset. For per-scene optimized evaluation, we fine-tune our pre-trained model for 1,500 steps. The quantitative results of these experiments are presented in Table \ref{tab:waymo_ft}. Notably, our method demonstrates superior performance not only compared to pose-free methods but also outperforms pose-given methods. These results confirm the effectiveness and versatility of our approach in achieving high-quality results across different datasets and evaluation scenarios.

\begin{table}[t]
    \caption{Quantitive results for novel view synthesis on Waymo \cite{waymo_open} dataset. Entries in \textbf{bold} indicate the best performance in a pose-free context, while \highlight{highlighted} represents the best overall.}
\centering
\begin{tabular}{cc|cc|ccc}
\toprule
\multirow{2}{*}{Metric} & \multirow{2}{*}{Scenes} & \multicolumn{2}{c|}{Pose-free \ding{56}}            & \multicolumn{3}{c}{Pose-free \ding{52}}  \\
                     &        & PixelSplat & IBRNet & Nope-NeRF~ & DBARF~ & Ours                                                                                                                                                                                                                                                                                                             \\ \midrule
\multirow{6}{*}{PSNR \(\uparrow\)}    & 003      & 33.04                          & 32.22  & 26.45      & 26.68  & \cellcolor[HTML]{FFFFC7}\textbf{33.22}                                                                                                                                                                                                                                                                                                              \\
                     & 019     & 32.77                          & 30.67  & 24.72      & 24.5   & \cellcolor[HTML]{FFFFC7}\textbf{33.21}                                                                                                                                                                                                                                                                                                              \\
                     & 036     & 33.94                          & 31.14  & 26.37      & 23.44  & \cellcolor[HTML]{FFFFC7}\textbf{34.12}                                                                                                                                                                                                                                                                                                             \\
                     & 069     & 31.8                           & 31.41  & 26.84      & 21.62  & \cellcolor[HTML]{FFFFC7}\textbf{33.51}                                                                                                                                                                                                                                                                                                             \\
                     & 081     & \cellcolor[HTML]{FFFFC7}32.55~ & 31.74  & 27.1       & 29.06  & \textbf{32.42} \\
                     & 126    & 36.38                          & 35.61  & 28.76      & 29.23  & \cellcolor[HTML]{FFFFC7}\textbf{36.61}                                                                                                                                                                                                                                                                                                             \\ \midrule
\multirow{6}{*}{SSIM \(\uparrow\)}    & 003      & 0.93                           & 0.923  & 0.781      & 0.859  & \cellcolor[HTML]{FFFFC7}\textbf{0.936}                                                                                                                                                                                                                                                                                                              \\
                     & 019     & 0.933                          & 0.907  & 0.766      & 0.829  & \cellcolor[HTML]{FFFFC7}\textbf{0.937}                                                                                                                                                                                                                                                                                                             \\
                     & 036     & 0.94                           & 0.91   & 0.792      & 0.818  & \cellcolor[HTML]{FFFFC7}\textbf{0.942}                                                                                                                                                                                                                                                                                                             \\
                     & 069     & 0.906                          & 0.896  & 0.775      & 0.762  & \cellcolor[HTML]{FFFFC7}\textbf{0.927}                                                                                                                                                                                                                                                                                                             \\
                     & 081     & \cellcolor[HTML]{FFFFC7}0.914~ & 0.904  & 0.759      & 0.876  & \textbf{0.904} \\
                     & 126    & 0.942                          & 0.937  & 0.801      & 0.881  & \cellcolor[HTML]{FFFFC7}\textbf{0.948}                                                                                                                                                                                                                                                                                                             \\  \midrule
\multirow{6}{*}{LPIPS \(\downarrow\)} & 003      & 0.129                          & 0.166  & 0.525      & 0.18   & \cellcolor[HTML]{FFFFC7}\textbf{0.083}                                                                                                                                                                                                                                                                                                              \\
                     & 019     & 0.135                          & 0.205  & 0.547      & 0.24   & \cellcolor[HTML]{FFFFC7}\textbf{0.079}                                                                                                                                                                                                                                                                                                             \\
                     & 036     & 0.159                          & 0.214  & 0.501      & 0.24   & \cellcolor[HTML]{FFFFC7}\textbf{0.096}                                                                                                                                                                                                                                                                                                             \\
                     & 069     & 0.132                          & 0.237  & 0.544      & 0.323  & \cellcolor[HTML]{FFFFC7}\textbf{0.121}                                                                                                                                                                                                                                                                                                             \\
                     & 081     & \cellcolor[HTML]{FFFFC7}0.131~ & 0.203  & 0.582      & 0.16   & \textbf{0.146} \\
                     & 126    & 0.09                           & 0.195  & 0.518      & 0.179  & \cellcolor[HTML]{FFFFC7}\textbf{0.080}           \\ \bottomrule
\end{tabular}
\label{tab:waymo_ft}

\end{table}

Among the pose-free methods, our approach significantly outperforms all other methods by a considerable margin. Specifically, in Scene 003, 019, 036, 069, 081, and 126, our method surpasses the competition by 6.54dB, 8.49dB, 7.75dB, 6.84dB, 3.36dB, and 7.38dB, respectively. The qualitative results presented in Figure \ref{fig:waymo_finetune} align with these findings, highlighting our method's superior ability to handle forward-facing scenarios compared to the other pose-free methods. This demonstrates the effectiveness and robustness of our approach in tackling challenging pose-free situations, resulting in significantly improved performance.

Among the pose-given methods, our method (still in pose-free) exhibits comparable or superior performance compared to pixelSplat in several scenes, namely Scene 003, 019, 036, 081, and 126. Additionally, our method outperforms pixelSplat by 1.71dB, achieving leading performance in Scene 069. It is worth noting that while pixelSplat may demonstrate better LPIPS performance in generalized settings (as shown in Table 1 on the main paper), our method surpasses it in the finetuning setting by significantly reducing the LPIPS metric. This result highlights the versatility of our method, as it not only performs well in generalized settings but also exhibits significant potential for improvement through finetuning.

\begin{figure}[H]
  \centering
  \includegraphics[width=12cm]{figures/ft_waymo.pdf}
  \caption{Qualitative results for novel view synthesis on Waymo~\cite{waymo_open} dataset with finetuning settings. We clip three images in each scene and compared our rendered results with other pose-free as well as pose-given methods. Correspond PSNR metric are listed below, discriminate areas are plot with 'O'.}
  \label{fig:waymo_finetune}
\end{figure}

\section{Depth Visualization}
\label{sec:depth_vis}
Figure \ref{fig:depth_vis} illustrates the visualization of depth results. Our proposed IPO-Net plays a vital role in estimating camera pose, showcasing its ability to accurately recognize geometric relationships and directly reflect the precision of the camera's pose estimation. Moreover, when compared to other pose-free methods like DBARF~\cite{chen2023dbarf} and IBRNet~\cite{wang2021ibrnet}, our proposed GGRt exhibits a clear advantage, yielding notably more accurate results in terms of depth estimation. This observation underscores the effectiveness and superiority of our method in capturing fine-grained depth information.

\begin{figure}[H]
  \centering
  \includegraphics[width=12cm]{figures/depth_vis.pdf}
  \caption{Qualitative results for novel view synthesis on Waymo~\cite{waymo_open} dataset with finetuning settings. }
  \label{fig:depth_vis}
\end{figure}

\section{Visualization}
Figure \ref{fig:depth_render_demo} shows the visualization results of selected clips from the rendered video. For the complete videos, please refer to the supplementary files.

\begin{figure}[H]
  \centering
  \includegraphics[width=12cm]{figures/depth_render_demo.pdf}
  \caption{Clips of rendered video in Waymo Scene 019, including rendered images and depth maps. }
  \label{fig:depth_render_demo}
\end{figure}

%
%
\bibliographystyle{splncs04}
\bibliography{egbib}